\documentclass[journal,twoside,web]{ieeecolor}
\usepackage{generic}
\usepackage{cite}
\usepackage{amsmath,amssymb,amsfonts}
\usepackage{graphicx}
\usepackage{algorithm}
\usepackage{algpseudocode}
\usepackage{hyperref}
\hypersetup{hidelinks=true}
\usepackage{textcomp}
\usepackage{makecell}
\usepackage{easyReview}
\usepackage{multirow}
\usepackage[acronym]{glossaries}

\newacronym{llm}{LLM}{Large Language Model}
\newacronym{rr}{RR}{Large Language Model}

\def\BibTeX{{\rm B\kern-.05em{\sc i\kern-.025em b}\kern-.08em
    T\kern-.1667em\lower.7ex\hbox{E}\kern-.125emX}}
\begin{document}


\title{\textbf{Nutri-ATLAS}: Embodied \textbf{A}gent for \textbf{T}abulated \textbf{L}ookup and \textbf{A}ssistance for \textbf{S}marter nutrition}


\author{Uttej~Kallakuri,
Boxun~Hu,
Ankur~A.~Butala,
Najim~Dehak, and Tinoosh~Mohsenin%
\thanks{Uttej~Kallakuri and Boxun~Hu contributed equally to this work.}%
\thanks{Uttej~Kallakuri, Boxun~Hu, Najim~Dehak, and Tinoosh~Mohsenin are with the Department of Electrical and Computer Engineering, Johns Hopkins University, Baltimore, MD 21218 USA.}%
\thanks{Ankur~A.~Butala, Najim~Dehak, and Tinoosh~Mohsenin are with the Geriatric Engineering at Johns Hopkins University, Baltimore, MD 21218 USA.}%
\thanks{Ankur~A.~Butala is with the Departments of Neurology and Neurosurgery, Johns Hopkins Medicine, Baltimore, MD 21287 USA.}%
\thanks{Corresponding author: Uttej~Kallakuri (e-mail: ukallak1@jhu.edu, bhu29@jh.edu).}
}



\maketitle

\begin{abstract}
Generative and Agentic IoT systems offer a promising foundation for digital healthcare applications that combine sensing, personalized reasoning, and autonomous interaction in real-world environments. Nutrition assistance is a natural use case, but existing \gls{llm} based systems are often limited to passive text interaction and static context, making them unreliable when food descriptions are ambiguous, or nutritional evidence is missing. To address these limitations, we propose Nutri-ATLAS, an Embodied Agent for Tabulated Lookup and Assistance for smarter nutrition in the real world. It integrates graph-grounded nutrition reasoning, hardware-aware LLM selection, and robot-based evidence acquisition. Nutri-ATLAS is not intended to replace clinical dietitians but to show how embodied agents can support nutrition by acquiring real world evidence. We frame nutrition assistance as a Generative and Agentic IoT problem in which an embodied assistant must reason over reported intake, dietary constraints, nutritional evidence, and foods physically available in the environment. Nutri-ATLAS builds a unified Food-Nutrient knowledge graph from USDA FoodData Central and FoodKG, learns 64-dimensional GATv2 food and recipe embeddings. A shared hybrid graph-text scoring supports food nutrition extraction, nutritional gap filling, substitute retrieval, and recipe-level meal composition, while an \gls{llm}-guided skill interface navigates landmarks, updates dietary-context and food-accessibility memory, and grounds recommendations in observed food availability. We evaluate Nutri-ATLAS across nutrient estimation, substitution retrieval, recipe recommendation, patient-profile adherence, edge deployment, and embodied execution in the real world. On HealthyFoodSubs, the hybrid retriever achieves 37.9\% MAP, 80.7\% RR@5, and 90.1\% RR@10. 
On NutriBench~v2, Dense+GAT retrieval grounds nutrient estimation across nine quantized Qwen3.5-9B configurations.
On PFoodReQ, Nutri-ATLAS reaches 78.8\% MAP, 83.0\% MAR, and 77.5\% F1. A patient-profile study shows adherence to allergy and healthy-target constraints for all selected cases.
\end{abstract}

\section{Introduction}
\label{sec:intro}

\begin{figure*}[ht]
  \centerline{\includegraphics[width=\textwidth]{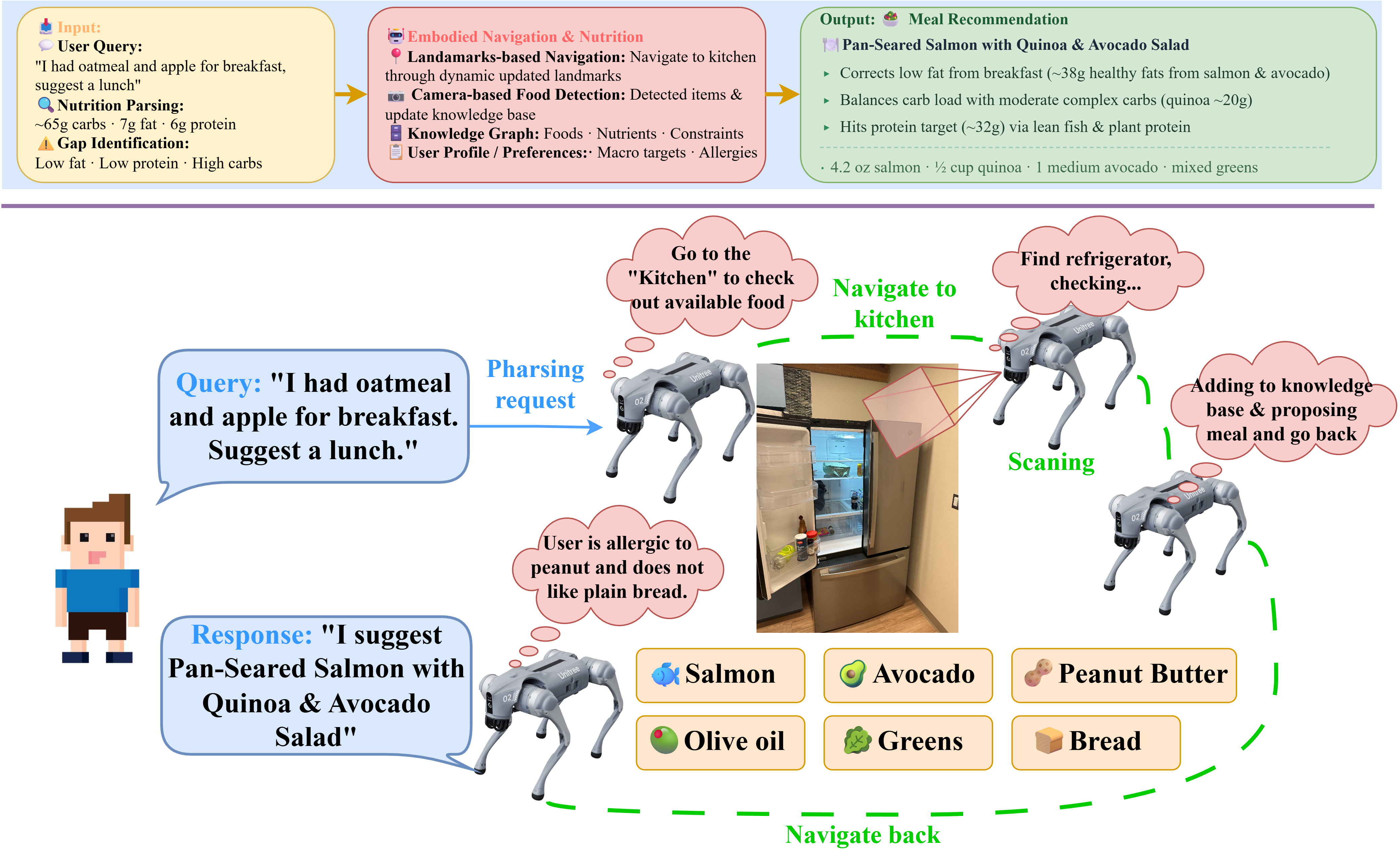}}
  \caption{\textbf{High-level overview of Nutri-ATLAS.} Given a user dietary query, \emph{e.g.}, ``I had oatmeal and apple for breakfast, suggest a lunch,'' the system parses the reported intake and identifies nutritional gaps with respect to the user's dietary preferences and restrictions. Nutritional reasoning is then coupled with embodied evidence acquisition: the robot navigates to the kitchen, inspects the refrigerator via camera-based food detection, and updates its knowledge base with newly observed food items and landmarks. Combining the user's reported intake, nutritional targets, and food present at hand, Nutri-ATLAS produces a grounded personalized recommendation, illustrated here with \emph{Pan-Seared Salmon with Quinoa \& Avocado Salad}, and can return feedback after completing the search.}
  \vspace{-10pt}
  \label{fig:high_level_system_overview}
\end{figure*}

What people eat governs a substantial share of preventable morbidity and mortality worldwide, with the 2017 Global Burden of Disease study estimating that dietary risks accounted for 11 million deaths and 255 million disability-adjusted life years globally in 2017~\cite{gbd2019dietaryrisks}. Sustained dietary improvement depends on individuals being able to track intake, understand nutritional gaps, and translate guidance into feasible meal choices. However, dietary monitoring remains difficult because self-reported recalls and food records are affected by random and systematic measurement errors, and manual food logging can be burdensome in daily use~\cite{subar2012asa24,gibson2017measurement,kirkpatrick2022measurement}. Recent mobile and AI-assisted dietary assessment tools reduce some friction, but they still face limitations in food recognition, portion estimation, database coverage, and nutrient inference~\cite{li2024nutritionapps,zheng2024aifoodnutrient,chotwanvirat2024foodimages,yan2025dietai24}. Large language model (LLM)-based assistants\cite{navardi2026metareasoning, kallakuri2026magrip, navardi2024metatinyml} can provide fluent nutrition explanations and meal suggestions, but clinical nutrition evaluations show that they remain vulnerable to incomplete evidence, hallucinated nutrient claims, and difficulty enforcing dietary constraints without external grounding~\cite{belkhouribchia2025llmclinicalnutrition}.

This paper addresses the following research problem: how to build a trustworthy nutrition assistant whose recommendations are faithful to structured nutritional knowledge and to foods actually observed in the user's environment, rather than only to the parametric tendencies of an LLM. Operationally, this requires the system to (i) ground user-reported intake in structured food-nutrient knowledge, (ii) generate recommendations from identified dietary gaps and user-specific dietary context, and (iii) actively acquire missing environmental evidence through embodied interaction when needed. In this work, \textit{personalization} refers to an explicit and persistent dietary context maintained by the system's Preference Memory: user-stated restrictions such as allergies or disease-related constraints, food preferences and dislikes, intake history, and prior offered/accepted recommendations. These records are used as hard constraint filters for safety-critical requirements, as soft re-ranking signals for preference-aware retrieval, and as conditioning context for LLM-based gap analysis and recommendation generation. This definition follows recent work on personalized nutrition and health-aware food recommendation, where useful meal guidance must jointly account for dietary requirements, health goals, preference signals, and practical feasibility~\cite{donovan2025personalizednutrition,wang2025personalizeddietreview,dong2026dietrelatedhrs,xu2024elcombo,kalpakoglou2025ainr}.

Grounding is foundational to this problem because both gap analysis and personalized recommendation depend on correctly identifying what the user consumed and what foods are available. Yet robust grounding is difficult for text-only retrieval: similarly named foods may differ substantially in nutrient composition, while nutritionally equivalent foods may share little vocabulary. We therefore argue that trustworthy nutrition assistance rests on two design principles: pairing structured graph-based food--nutrient representations with semantic text understanding, and integrating this retrieval pipeline with an embodied action interface that can acquire missing environmental evidence. Recent LLM-based nutrition systems, knowledge-graph dietary assistants, and closed-loop meal-planning agents point toward this direction, but most remain limited to text, images, or static user profiles rather than physically grounded food availability~\cite{khamesian2025nutrigen,gao2025healthgenie,xu2026closedloopnutrition,wang2026heirag}.

To this end, we propose \textbf{Nutri-ATLAS}, whereby ATLAS reflects the system's core function in structured nutritional lookup: \textbf{\textit{A}}gent for \textbf{\textit{T}}abulated \textbf{\textit{L}}ookup and \textbf{\textit{A}}ssistance for \textbf{\textit{S}}marter nutrition. Nutri-ATLAS unifies two complementary representations of food-nutrition-aware graph embeddings learned by a Graph Attention Network and semantic text embeddings from a pretrained text encoder-and fuses them through a hybrid retrieval primitive that drives a four-stage nutrition-aware pipeline spanning food grounding, gap-aware recommendation, substitute retrieval, and meal-level composition. The retrieval and recommendation pipeline incorporates the Preference Memory to enforce user restrictions, bias candidates toward learned preferences, and condition final explanations on dietary context. It is paired with an LLM-guided robotic skill interface that acquires missing environmental evidence, updates the Food Accessibility store, and feeds observed availability back into recommendation, yielding the end-to-end embodied workflow illustrated in Fig.~\ref{fig:high_level_system_overview}. For resource-constrained edge deployment, Nutri-ATLAS further selects among quantized LLM backbones through a unified accuracy score and Pareto analysis over nutrition accuracy, generation throughput, and energy efficiency~\cite{shaharear2024vit, mazumder2023reg, mazumder2024reg, rashid2}. It is important to distinguish the two complementary roles in Nutri-ATLAS. The robotic component localizes the user's food environment by navigating to landmarks, inspecting visible items, and updating the Food Accessibility store. Personalization, by contrast, resides in the retrieval and recommendation pipeline through Preference Memory, which stores dietary restrictions, preferences, and prior recommendation feedback. The robot does not independently deliver nutrition advice; instead, its observations are returned to the graph-grounded recommendation pipeline, where they are combined with nutritional evidence and user-specific dietary context to produce the final recommendation. The key contributions of this work are summarized as follows:
\begin{itemize}

    \item \textbf{Unified food-nutrition graph and text embeddings.} We construct a unified nutrition knowledge base from USDA FoodData Central and FoodKG, and train a two-layer Graph Attention Network (GATv2) that jointly learns 64-dimensional embeddings for 9{,}991 foods and 82{,}238 recipes through heterogeneous food--nutrient--recipe--tag message passing. In parallel, food and recipe descriptions are encoded by a pretrained text encoder into a complementary 1{,}024-dimensional semantic space. The two spaces capture orthogonal similarity signals: graph embeddings encode nutritional and ingredient structure invisible to text alone, while text embeddings capture lexical and cross-language name similarity.

    \item \textbf{Personalized nutrition-aware retrieval and food recommendation.} We introduce a four-stage nutrition-aware retrieval pipeline - food mention grounding, nutritional gap filling, equivalent substitute retrieval, and meal-level recipe composition - unified by a single hybrid scoring primitive that fuses graph and text embeddings at every stage. User dietary preferences and restrictions are incorporated as soft preference re-ranking signals and hard constraint filters, enabling nutrition-grounded recommendations that adapt to individual needs without model retraining.

    \item \textbf{Hardware-aware LLM selection for nutrition reasoning and navigation.} We present a deployment-oriented framework for selecting LLMs for nutrition reasoning under edge-resource constraints. Under a fixed Dense + GAT retrieval pipeline, we evaluate multiple quantized Qwen3.5-9B variants not only by nutrient-estimation quality across carbohydrate, protein, fat, and energy, but also by on-device execution behavior, including weighted generation throughput, tail latency, and energy per generated token on the target hardware platform. To support practical deployment, we derive a unified task-quality score that combines graph-based substitution retrieval, personalized food recommendation, and NutriBench performance, and then perform multi-objective hardware-aware model selection using Pareto-front analysis to identify ideal deployment candidates.

    \item \textbf{Integrated embodied nutrition system in the real world.} We present Nutri-ATLAS as an integrated system that couples the graph-grounded nutrition pipeline with an LLM-guided robotic skill interface for active environmental evidence acquisition. The LLM planner invokes navigation, landmark listing, object querying, preference updating, and food-accessibility skills to close the loop between dietary reasoning and environmental perception, progressively expanding its knowledge base over long-horizon interactions and feeding confirmed food availability back into recommendation.

\end{itemize}


\begin{figure*}[ht]
\centerline{\includegraphics[width=\textwidth]{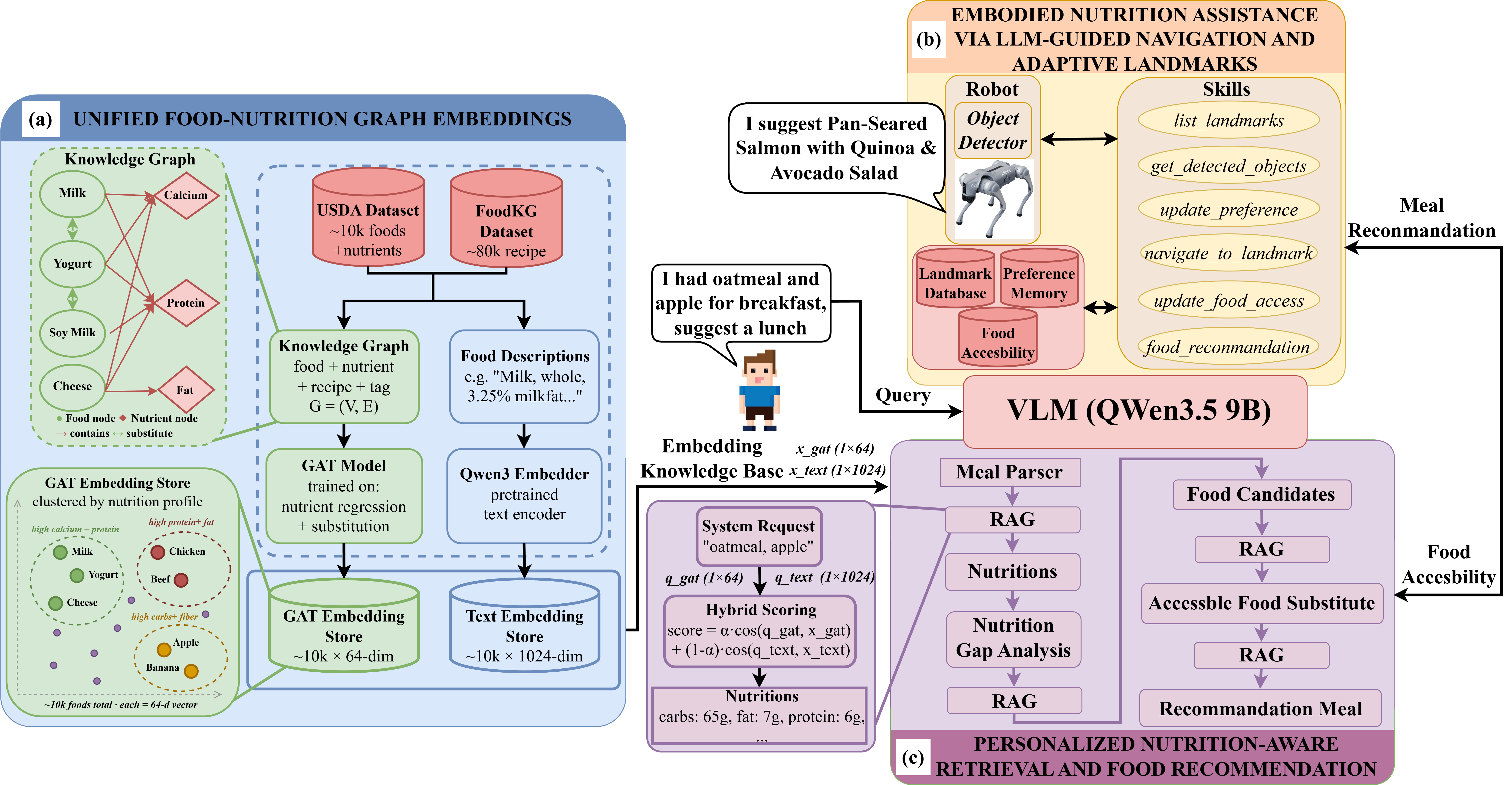}}
\caption{\textbf{System architecture and technical pipeline of Nutri-ATLAS.}
(a) \textit{Unified embedding module.} A heterogeneous graph $G=(V,E)$ built from USDA foods and nutrients and FoodKG recipes is encoded by a GATv2 into nutrition-aware embeddings $x_{\mathrm{gat}}\in\mathbb{R}^{64}$, while a pretrained text embedder encodes the corresponding descriptions into $x_{\mathrm{text}}\in\mathbb{R}^{1024}$.
(b) \textit{Embodied assistance module.} An LLM planner drives a robotic skill library --- landmark listing, object querying, preference and food-access updating, and recommendation --- over three persistent stores (a landmark database, preference memory, and food-accessibility memory), acquiring missing environmental evidence and progressively updating its knowledge base.
(c) \textit{Retrieval and recommendation module.} A user query is encoded into $(q_{\mathrm{text}},q_{\mathrm{gat}})$ and ranked against candidates by the shared hybrid primitive $s=\alpha\cos(q_{\mathrm{gat}},x_{\mathrm{gat}})+(1-\alpha)\cos(q_{\mathrm{text}},x_{\mathrm{text}})$. The pipeline grounds reported intake to nutrient profiles, analyzes nutritional gaps, and retrieves gap-filling foods, substitutes, and meals; food-accessibility evidence from (b) is fed back to ground recommendations.}
\vspace{-10pt}
\label{fig:system_overview}
\end{figure*}
\vspace{-2ex}
\section{Related Works}
\label{sec:related_work}

\subsection{LLM-Based Nutrition Assessment and Dietary Assistance}
Recent work has explored large language models (LLMs) as tools for dietary assessment, meal planning, and nutrition education, but also highlights the need for grounding, constraint checking, and clinical oversight. NutriBench provides one of the first large-scale benchmarks for estimating macronutrients from natural-language meal descriptions and shows that LLMs can support nutrition estimation, but remain sensitive to prompting and external evidence quality~\cite{dhaliwal2025nutribench}. Complementary evaluations using the Registered Dietitian exam show that leading LLMs can answer many nutrition questions accurately, but performance varies with prompting strategy, difficulty level, and domain, motivating careful validation rather than direct clinical substitution~\cite{azimi2025llmnutritionrd}. Similarly, studies of LLM-generated diabetes meal plans and image-based nutrition estimation report clinically relevant deviations, systematic underestimation, and portion-size errors, indicating that general-purpose models should be treated as assistive tools rather than autonomous nutritionists~\cite{karakas2026clinicalnutritionllm,fridolfsson2025nutritionimages}. More recent multimodal work, such as NutriMLLM, further demonstrates both the promise and limitations of vision-language models for detailed nutrient estimation from food images~\cite{yan2026nutrimllm}.

Several systems move from assessment toward recommendation. ChatDiet uses an LLM-augmented framework for interactive nutrition-oriented food recommendation, while recent Healthy Eating Index-informed RAG methods anchor recommendation in standardized dietary-quality metrics~\cite{yang2024chatdiet,wang2026heirag}. These efforts demonstrate the value of LLMs for explanation and interaction, but they typically operate over text, profiles, or static databases. In contrast, Nutri-ATLAS frames nutrition assistance as an embodied, evidence-acquisition problem: the system grounds reported intake in structured food--nutrient knowledge, checks dietary context and restrictions, and updates recommendations using foods observed in the user's environment.
\subsection{Food Knowledge Graphs, Substitution, and Recipe Recommendation}
Knowledge graphs have become an important foundation for food recommendation because they can encode ingredients, recipes, nutrients, tags, and health-related constraints in a structured form. FoodKG introduced a semantics-driven food knowledge graph that links recipes, food entities, and nutritional information for food recommendation~\cite{haussmann2019foodkg}. PFoodReQ later formulated personalized food recommendation as constrained question answering over a large-scale food knowledge graph, explicitly modeling user requirements such as dietary preferences, dislikes, health guidelines, and ingredient constraints~\cite{chen2021pfoodreq}. These works provide strong foundations for graph-based food reasoning and are directly related to Nutri-ATLAS's use of FoodKG-derived recipes and constrained recommendation.

Food substitution has also been studied through knowledge graph embeddings and graph neural networks. Loesch et al. introduced a food substitution dataset and evaluated knowledge graph embeddings for identifying nutritionally similar alternatives~\cite{loesch2022automated}. Their later HealthyFoodSubs work uses GraphSAGE and GAT models with Nutri-Score ranking to identify healthier substitutes within food categories~\cite{loesch2024automated}. Other recent food recommendation systems, such as nutrition-related knowledge graph neural networks, incorporate nutrient structure into graph-based recommendation to reduce purely preference-driven repetition and improve nutritional diversity~\cite{ma2024nrkgnn}. NGQA extends this direction by introducing a nutritional graph question-answering benchmark for health-aware dietary reasoning over user-specific contexts~\cite{zhang2025ngqa}. Nutri-ATLAS differs from these works by unifying substitution retrieval, nutrient grounding, gap filling, and meal composition under a shared graph--text retrieval primitive, and by coupling this static knowledge with food-availability evidence.

\subsection{Retrieval-Augmented and Graph-Grounded Generation}
Retrieval-augmented generation (RAG) reduces hallucination by conditioning LLM outputs on external evidence rather than relying only on parametric knowledge~\cite{lewis2020rag, mao2025multi}. Traditional lexical retrieval, including BM25, remains useful for matching explicit terms, but it is limited when nutritional similarity is not reflected in surface vocabulary~\cite{robertson2009bm25}. Dense text retrieval improves semantic matching, yet it can still retrieve textually similar but nutritionally mismatched foods. GraphRAG-style methods address related limitations by organizing retrieved information through entities, relationships, and graph summaries, demonstrating the value of structured retrieval for complex queries~\cite{edge2024graphrag}. Recent embedding models such as Qwen3-Embedding further improve multilingual and semantic retrieval capabilities, making them useful for free-text food mention grounding and recipe retrieval~\cite{zhang2025qwen3embedding}.

Nutri-ATLAS builds on this retrieval literature but applies it to a different problem setting. Rather than constructing a graph over text documents, it learns nutrition-aware GATv2 embeddings over a heterogeneous food--nutrient--recipe graph and fuses them with semantic text embeddings. This combination lets the system retrieve foods that are textually related, nutritionally similar, or both. The resulting hybrid retrieval primitive is shared across food nutrition extraction, nutritional gap filling, substitute retrieval, and recipe-level meal composition, allowing LLM generation to be grounded in structured nutrient evidence rather than unsupported free-form reasoning.

\subsection{Embodied Agents, Memory, and Environmental Evidence}
Embodied LLM and multimodal-agent research has increasingly focused on whether language models can perceive, plan, and act in physical or simulated environments. EmbodiedBench evaluates multimodal LLMs across vision-driven embodied tasks and shows that current models still struggle with long-horizon planning, spatial reasoning, and low-level action execution~\cite{yang2025embodiedbench, miki-atlasv2-aaai-sss-2025, uttej-atlas}. EmbodiedEval similarly evaluates multimodal LLMs as embodied agents across navigation, interaction, and spatial question-answering tasks, emphasizing that static image or text benchmarks do not fully capture embodied capability~\cite{cheng2025embodiedeval}. Work on long-horizon memory, such as MemGPT, further shows the importance of external memory mechanisms for agents that must maintain state across extended interactions~\cite{packer2023memgpt}.

Nutri-ATLAS adopts a constrained and task-specific embodiment strategy. The robot is not treated as a general-purpose nutrition expert; instead, it provides environmental grounding by navigating to landmarks, inspecting food items, updating food-accessibility memory, and feeding observed availability back into the recommendation pipeline. This separation clarifies the role of embodiment: dietary reasoning and constraint-aware recommendation remain in the graph-grounded retrieval pipeline, while the robot acquires missing physical evidence needed to make recommendations actionable in the user's current environment.

\vspace{-4ex}
\subsection{Efficient LLM Deployment and Hardware-Aware Model Selection}
Deploying LLM-based assistants on embodied or edge platforms requires attention to model size, memory pressure, latency, throughput, and energy. Structured compression methods such as LLM-Pruner, SliceGPT, MaGrIP~\cite{kallakuri2026magrip, kallakuri2024resource} reduce model footprint and inference cost by pruning coupled structures or replacing large weight matrices with smaller dense representations~\cite{ma2023llmpruner,ashkboos2024slicegpt, navardi2025e2ar, walczak2025bitmedvit, aalishah2025medmambalite}. These methods address the general problem of making LLMs more efficient, but they do not directly decide which compressed or quantized model is best for a specific embodied nutrition task.

Nutri-ATLAS therefore treats deployment as a task- and hardware-aware selection problem rather than only a compression problem. Under a fixed Dense+GAT retrieval pipeline, quantized Qwen3.5-9B variants are compared using nutrient-estimation accuracy, unified task-quality scoring, generation throughput, tail latency, and energy per generated token. Pareto analysis then exposes operating points for accuracy-first, energy-first, performance-first, or balanced deployment. This complements model-compression literature by connecting LLM selection directly to nutrition reasoning quality and measured Jetson Orin Nano execution behavior. Overall, prior work has made important progress in LLM-based nutrition estimation, food knowledge graphs, constrained food recommendation, graph-based substitution, RAG, embodied-agent evaluation, and efficient LLM deployment. However, these directions are typically studied in isolation. Nutrition LLMs often lack structured grounding and physical context; food knowledge graph systems generally remain offline or database-centric; embodied-agent benchmarks evaluate planning but not nutrition-specific reasoning; and compression work rarely connects model choice to task-level nutrition quality. Nutri-ATLAS addresses this gap by integrating graph--text nutrition retrieval, constraint-aware meal recommendation, hardware-aware LLM selection, and robot-based evidence acquisition into a single closed-loop embodied nutrition assistance system.

\section{Proposed Approach: Nutri-Atlas}
\label{sec:approach}


\subsection{Problem Setup: Embodied Agentic Nutrition Assistance}
\label{subsec:problem_setup}

Given a query $q$ that reports recent intake and requests a recommendation - e.g., ``I had an apple and milk for breakfast, what should I eat for lunch?'' or ``Suggest a high-fiber snack I can grab from the kitchen'' - the agent must return a recommendation that is nutritionally appropriate, preference-aware, and grounded in the user's available foods. When the KB is insufficient, the agent invokes embodied skills - navigating to landmarks, querying detected objects, and updating food accessibility - to gather evidence before responding. The following subsections detail the three modules of Fig.~\ref{fig:system_overview}: the unified food-nutrition embeddings (Sec.~\ref{subsec:graph_kb}), the retrieval and recommendation pipeline (Sec.~\ref{subsec:recommendation}), and the embodied assistance module (Sec.~\ref{subsec:embodied_system}), along with hardware-aware model selection.

\begin{figure}[!t]
\centering
\includegraphics[width=\columnwidth]{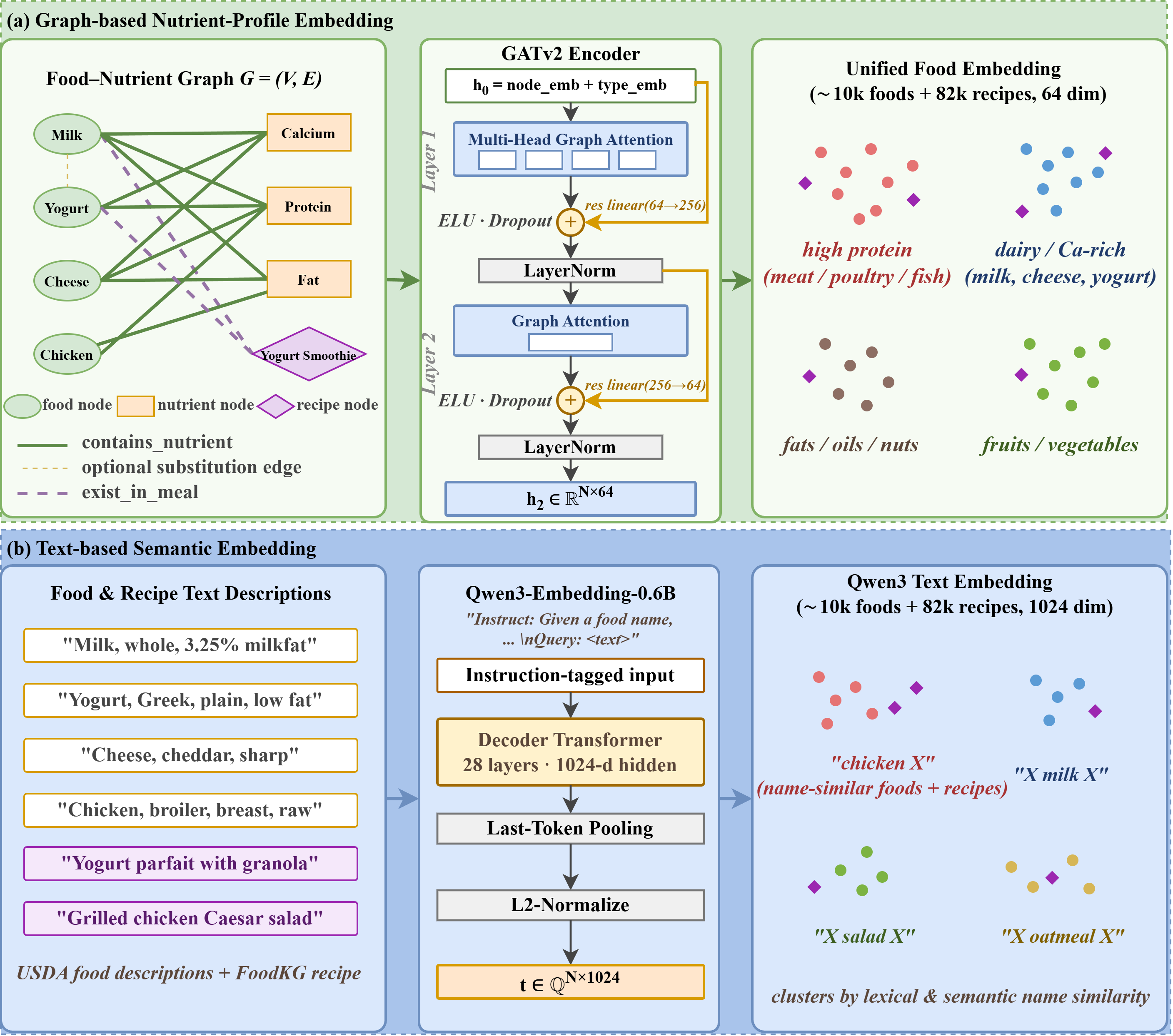}
\caption{\textbf{Nutri-ATLAS unified food-nutrient embeddings}, expanding the embedding module of Fig.~\ref{fig:system_overview}(a) and built from USDA FoodData Central and FoodKG. \textbf{(a) Graph branch:} a two-layer GATv2 over the heterogeneous food--nutrient--recipe--tag graph $G=(V,E)$ maps each node to a 64-d embedding; foods cluster by nutrient profile (high-protein, dairy, fats, fruits/vegetables). \textbf{(b) Text branch:} Qwen3-Embedding-0.6B encodes food/recipe descriptions into 1024-d embeddings that cluster by lexical and semantic name similarity.}

\label{fig:food_nutrient_embeddings}
\vspace{-8pt}
\end{figure}

\subsection{Unified food–nutrition graph embeddings}
\label{subsec:graph_kb}

\begin{figure*}[t]
  \centering
  \includegraphics[width=\textwidth]{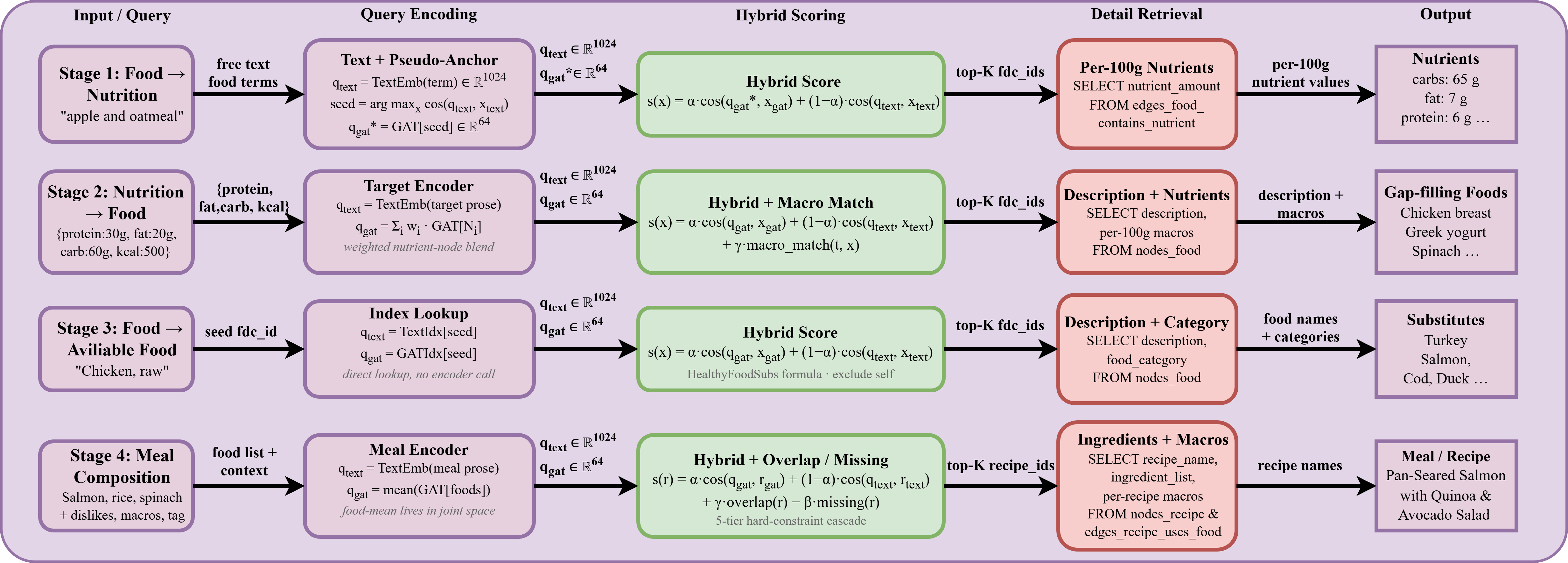}
    \caption{\textbf{Nutri-ATLAS personalized nutrition-aware retrieval and recommendation pipeline}, operating over USDA FoodData Central foods and nutrients and the FoodKG recipe corpus.
    Four RAG stages share one hybrid scoring primitive
    $s(\cdot)=\alpha\cos(\mathbf{q}_\text{gat},\mathbf{x}_\text{gat})+(1-\alpha)\cos(\mathbf{q}_\text{text},\mathbf{x}_\text{text})$
    over two stores: an \textit{Embedding Index} (food and recipe text $\in\mathbb{R}^{1024}$, GAT $\in\mathbb{R}^{64}$)
    for cosine ranking, and a \textit{Knowledge Base} (DuckDB; 9{,}991 foods, 626 nutrients, 82{,}238 recipes)
    for SQL pre-filtering and structured detail retrieval.
    \textbf{Stage~1} (\textit{Food$\to$Nutrition}) grounds free-text food mentions to USDA entries via a
    text-bootstrapped pseudo-anchor $\mathbf{q}_\text{gat}^{*}$.
    \textbf{Stage~2} (\textit{Nutrition$\to$Food}) retrieves gap-filling foods, encoding the LLM-derived target
    as a magnitude-weighted blend of nutrient-node GAT embeddings plus a macro-match term.
    \textbf{Stage~3} (\textit{Food$\to$Similar Food}) finds nutritionally equivalent substitutes by direct
    index lookup on the seed food's precomputed vectors.
    \textbf{Stage~4} (\textit{Meal Composition}) retrieves recipes from the FoodKG corpus using a mean-GAT query
    in the shared food--recipe space, re-ranked by ingredient-overlap and pantry-availability terms under a
    five-tier hard-constraint relaxation cascade.}
  \label{fig:rag_pipeline}
\end{figure*}

Nutri-ATLAS represents each food and recipe in two complementary embedding spaces (Fig.~\ref{fig:food_nutrient_embeddings}). A \textit{graph branch} encodes nutritional structure with a GATv2 encoder over a heterogeneous food-nutrient graph, producing 64-dimensional embeddings that reflect nutrient composition, ingredient co-occurrence, and substitution relationships. A \textit{text branch} encodes food and recipe descriptions with Qwen3-Embedding-0.6B into 1024-dimensional embeddings capturing lexical and semantic description similarity~\cite{zhang2025qwen3embedding}. The two are complementary: graph embeddings surface nutritionally similar foods that share no vocabulary, while text embeddings ground free-text mentions regardless of nutrient content.

\textbf{Heterogeneous nutrition graph.} $G=(V,E)$ has four node types - 9{,}991 food, 626 nutrient, 82{,}238 recipe, and 238 tag nodes ($|V|=93{,}093$) - and four edge types. The food--nutrient portion is constructed from USDA FoodData Central~\cite{fukagawa2022fooddata,usda2024fooddatacentral}, while the recipe and ingredient relations are derived from FoodKG~\cite{haussmann2019foodkg}. Food$\leftrightarrow$Nutrient edges store a $\log(1+\texttt{amount})$ attribute that compresses the heavy-tailed nutrient scale; Recipe$\leftrightarrow$Food edges encode ingredient containment; Recipe$\leftrightarrow$Tag edges assign meal-type and cuisine labels; and substitution edges link food pairs from HealthyFoodSubs~\cite{loesch2024automated}. Only the Food$\leftrightarrow$Nutrient and substitution relations are supervised; recipe and tag embeddings emerge through heterogeneous message passing~\cite{gilmer2017neural}.

\textbf{GATv2 encoder.} A two-layer GATv2 maps $G$ to 64-dimensional node embeddings. We use GATv2 because the nutrients that drive similarity vary by food category, and its attention conditions jointly on source and target nodes, giving more expressive aggregation than the original GAT formulation~\cite{velickovic2018gat,brody2022gatv2}. Each node is initialized as the sum of a learned node embedding and a node-type embedding. The first layer uses 4 attention heads of width $d_h=64$ and the second a single head back to $\mathbb{R}^{64}$; both apply ELU~\cite{clevert2016elu}, dropout, a width-matched residual projection, and LayerNorm~\cite{ba2016layernorm}.

The encoder is trained with a multi-task objective on observed Food$\rightarrow$Nutrient edges:
\begin{equation}
\label{eq:gat_loss}
\mathcal{L}=\mathcal{L}_{\mathrm{amt}}+\lambda_{\mathrm{exist}}\mathcal{L}_{\mathrm{exist}}+\mathcal{L}_{\mathrm{subs}},
\end{equation}
where $\mathcal{L}*{\mathrm{amt}}$ is a Smooth~L1 loss~\cite{girshick2015fastrcnn} on standardized $\log(1+\texttt{amount})$ targets, $\mathcal{L}*{\mathrm{exist}}$ is a binary cross-entropy loss for nutrient-edge existence using sampled non-edges as negatives ($\lambda_{\mathrm{exist}}=0.4$), and $\mathcal{L}_{\mathrm{subs}}$ is a binary cross-entropy loss for food substitution. We train for 90 epochs with AdamW~\cite{loshchilov2019adamw} ($\eta=2\times10^{-3}$, weight decay $10^{-4}$) and a ReduceLROnPlateau schedule (factor 0.6, patience 4).

\textbf{Text branch.} In parallel, Qwen3-Embedding-0.6B~\cite{zhang2025qwen3} - a 28-layer decoder-only transformer with last-token pooling - encodes USDA food descriptions and FoodKG recipe names as documents into a shared 1024-dimensional space, yielding a $9{,}991\times1024$ food index and an $82{,}238\times1024$ recipe index, $\ell_2$-normalized for cosine search. Following the model's query-document convention, a free-text food mention is instead encoded at query time with an instruction prefix, focusing similarity on nutritional relevance rather than surface word overlap.

\textbf{Hybrid scoring.} At inference, the two spaces are fused by a single primitive shared across all four retrieval stages (Sec.~\ref{subsec:recommendation}):
\begin{equation}
\label{eq:hybrid_score}
s(x) = \alpha\,\cos(\mathbf{q}_{\mathrm{gat}},\mathbf{x}_{\mathrm{gat}}) + (1-\alpha)\,\cos(\mathbf{q}_{\mathrm{text}},\mathbf{x}_{\mathrm{text}}),
\end{equation}
with $\alpha=0.5$. When the query is free text rather than a graph node, a \textit{pseudo-anchor} supplies the graph query: the text top-1 match's precomputed GAT embedding becomes $\mathbf{q}_{\mathrm{gat}}^{*}$, enabling fusion without a graph-inference call at query time.

\subsection{Personalized Nutrition-Aware Retrieval and Food Recommendation}
\label{subsec:recommendation}

Given a user query reporting recent intake, Nutri-ATLAS generates personalized recommendations through a four-stage RAG pipeline interleaved with two LLM calls (Fig.~\ref{fig:rag_pipeline}). Every stage applies the hybrid primitive of Eq.~(\ref{eq:hybrid_score}) over the food index or the recipe index; the Embedding Index serves cosine ranking, while the DuckDB Knowledge Base handles SQL pre-filtering and structured detail lookup~\cite{raasveldt2019duckdb}. This combines retrieval-augmented generation~\cite{lewis2020rag}, dense semantic retrieval~\cite{karpukhin2020dpr,zhang2025qwen3embedding}, and structured database lookup.

The pipeline opens with an LLM-driven gap analysis. From the meal description $q$, the grounded nutrient profile $\mathcal{N}_{\text{intake}}$, and the preference profile $\mathcal{P}_{\text{user}}$, the first LLM call determines which macro- and micronutrients are underrepresented and emits a structured target for the next meal:
\begin{equation}
\text{LLM}_1(q,\, \mathcal{N}_{\text{intake}},\, \mathcal{P}_{\text{user}}) \rightarrow \{\texttt{reasoning},\, \texttt{targets}\}.
\end{equation}

\textbf{Stage 1 (Food$\to$Nutrition)} grounds the food mentions in $q$ to USDA entries to populate $\mathcal{N}_{\text{intake}}$. As the query is free text rather than a graph node, a pseudo-anchor supplies the graph side: the text top-1 match yields $\mathbf{q}_{\mathrm{gat}}^{*}$, enabling hybrid scoring over all 9{,}991 foods without a graph-inference call. Per-100g nutrients for the top-$K$ matches are then fetched from the Knowledge Base.

\textbf{Stage 2 (Nutrition$\to$Food)} retrieves gap-filling foods for the target. The target is encoded into both spaces - a prose macronutrient descriptor gives $\mathbf{q}_{\mathrm{text}}$, and a magnitude-weighted blend of nutrient-node GAT embeddings gives $\mathbf{q}_{\mathrm{gat}}$. Over a SQL-filtered pool of meal-appropriate foods, candidates are ranked by the hybrid score plus a structured macro-match term:
\begin{equation}
\begin{split}
s(x) = &\;\alpha\,\cos(\mathbf{q}_{\mathrm{gat}},\,\mathbf{x}_{\mathrm{gat}}) + (1-\alpha)\,\cos(\mathbf{q}_{\mathrm{text}},\,\mathbf{x}_{\mathrm{text}}) \\
       &+ \gamma\,\texttt{macro\_match}(\texttt{targets},\, x),
\end{split}
\end{equation}
with $\gamma=0.5$.

\textbf{Stage 3 (Food$\to$Similar Food)} expands the set with nutritionally equivalent substitutes. Since each recommended food is already a graph node, both query vectors come from direct index lookup, and candidates are hybrid-ranked over the available-food pool (or the full index when availability is unknown), excluding the seed. This surfaces alternatives that share no name overlap with the original yet match it nutritionally.

\textbf{Stage 4 (Meal Composition)} lifts individual foods to concrete recipes. The recommended foods are pooled into a mean GAT query $\mathbf{q}_{\mathrm{gat}} = \mathrm{mean}(\mathbf{g}_i)$, valid because recipe and food GAT vectors share one jointly trained space and recipes cluster near their constituent ingredients. A meal-type tag narrows the corpus to the $\sim$200 candidate recipes, which are hybrid-ranked and then re-scored with ingredient-overlap and pantry-missing terms; a five-tier relaxation cascade progressively drops the macro, overlap, and finally dislike constraints so that a non-empty, dietary-compliant result is always returned.

\textbf{Personalization and generation.} A closed-loop tracker logs each recommended food as offered and, if selected, accepted, yielding a preference score
\begin{equation}
p_i = \frac{\text{chosen}_i}{\text{offered}_i},
\end{equation}
which defaults to a neutral $0.5$ for foods not yet offered. These scores re-rank candidates before the second LLM call, which composes the gap-analysis reasoning, the ranked foods (Stages~2-3), the top recipes (Stage~4), and the preference summary into a natural-language recommendation that is nutritionally appropriate, preference-aware, and grounded in available foods~\cite{hu2008implicit,rendle2009bpr}.

\begin{figure*}[t]
  \centering
  \includegraphics[width=\textwidth]{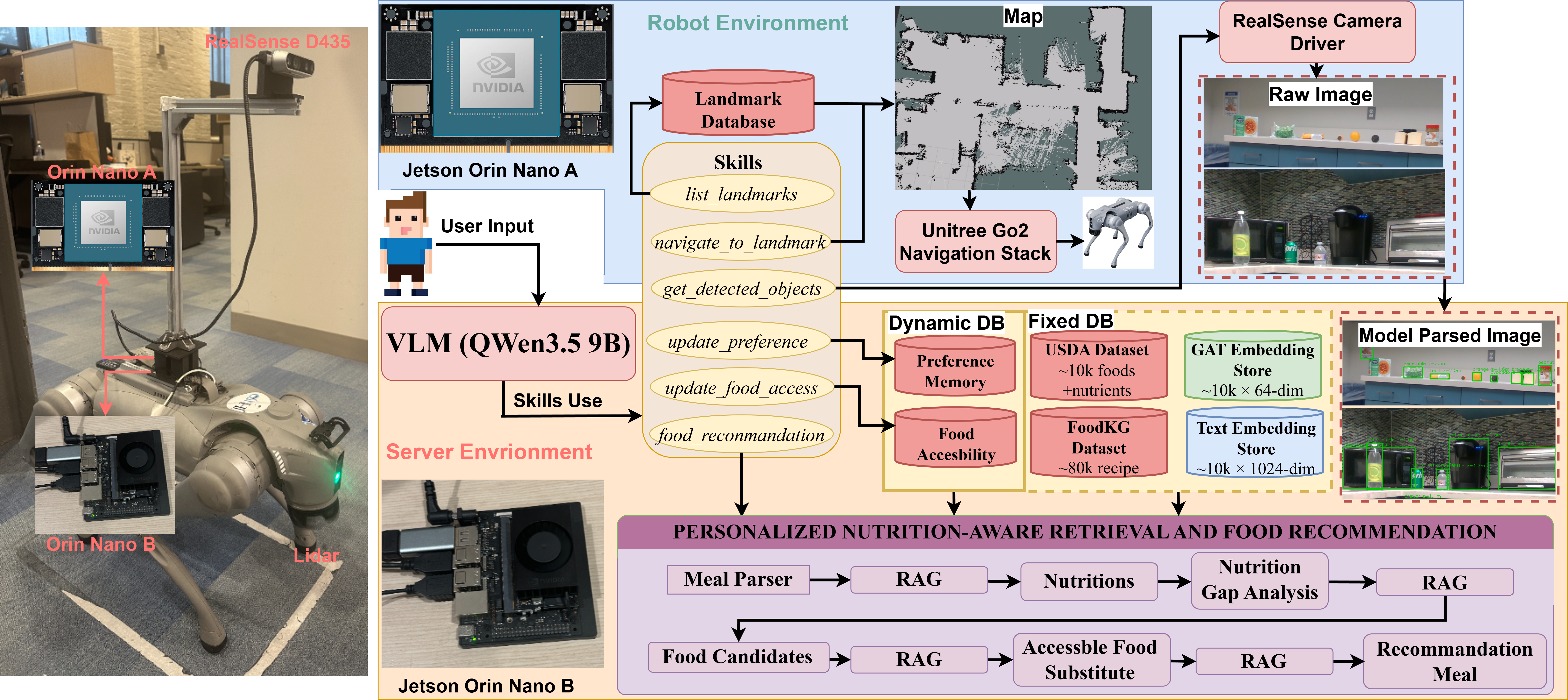}
    \caption{\textbf{Nutri-ATLAS real-world hardware and system integration setup. }The embodied assistant is deployed on a Unitree Go2 robot equipped with a RealSense-D435 camera and onboard sensing, with two Jetson Orin Nano modules (A and B) supporting the robot-side and server-side execution. A Qwen3.5-9B-based planner receives the user request and invokes typed skills for landmark listing, navigation, object querying, preference updating, food-access updating, and nutrition recommendation. The robot navigation stack uses the environment map and landmark database to move through the physical space, while the RealSense camera stream is parsed into detected food objects that update the dynamic food-accessibility store. These observations are combined with persistent preference memory, fixed USDA/FoodKG databases, and precomputed GAT and text embedding stores to drive the personalized nutrition-aware retrieval pipeline, producing recommendations grounded in both structured nutrition knowledge and foods observed in the environment.}
\label{fig:hardware_setup}

  \label{fig:realword-hardware}
\end{figure*}

\subsection{Hardware-aware multi-objective pareto analysis}
\label{subsec:model_selection}

To support deployment on resource-constrained embodied platforms, Nutri-ATLAS includes a hardware-aware model evaluation and selection module. The goal is to select a quantized LLM configuration that preserves nutrition reasoning quality while reducing inference cost. Rather than selecting a model only by nutrient-estimation accuracy, we formulate deployment as a multi-objective selection problem involving nutrition accuracy, generation throughput, and energy efficiency~\cite{deb2001multiobjective,emmerich2018tutorial}.

For each quantized model $m$, we first evaluate nutrient-estimation accuracy on NutriBench~v2 under the fixed Dense + GAT retrieval setting. Let $A_n(m)$ denote the accuracy for nutrient category $n\in\{\mathrm{Carbohydrate},\mathrm{Protein},\mathrm{Fat},\mathrm{Energy}\}$. We compute the model-level nutrition score using a weighted geometric mean:
\begin{equation}
S_{\mathrm{model}}(m)=\prod_{n}\left(\frac{A_n(m)}{100}\right)^{w_n},
\end{equation}
where $w_n$ is the importance weight of nutrient category $n$ and $\sum_n w_n=1$. The geometric mean penalizes models that perform poorly on any single nutrient category, preventing model selection from being dominated by one strong metric.

To summarize end-to-end nutrition capability, we combine model-level nutrient estimation with graph-based substitution retrieval and personalized recommendation quality. Let $S_{\mathrm{graph}}$ denote the food-substitution retrieval score, $S_{\mathrm{rec}}$ denote the personalized recommendation score, and $S_{\mathrm{model}}(m)$ denote the NutriBench score of model $m$. The Unified Accuracy Score is defined as
\begin{equation}
S_{\mathrm{unified}}(m)=100\cdot S_{\mathrm{graph}}^{w_g}S_{\mathrm{rec}}^{w_r}S_{\mathrm{model}}(m)^{w_m},
\label{eq:unified_acc}
\end{equation}
where $w_g+w_r+w_m=1$. This score provides a task-oriented summary for comparing quantized models while preserving the contributions of graph substitution, recommendation quality, and nutrient estimation.

For deployment-oriented selection, we further perform Pareto analysis over three objectives: maximizing accuracy, maximizing weighted generation throughput, and minimizing energy per generated token. For two models $m_i$ and $m_j$, $m_i$ dominates $m_j$ if
\begin{equation}
A_i\geq A_j,\quad T_i\geq T_j,\quad E_i\leq E_j,
\end{equation}
with at least one strict inequality, where $A$ is accuracy, $T$ is weighted generation throughput, and $E$ is energy per generated token. The Pareto frontier contains models that are not dominated by any other configuration, enabling Nutri-ATLAS to choose different deployment modes such as energy-first, accuracy-first, performance-first, and balanced operation.

\subsection{Embodied Nutrition Assistance and System Integration}
\label{subsec:embodied_system}

Fig.~\ref{fig:hardware_setup} shows the deployed system: a Unitree Go2 carrying a RealSense~D435 and two Jetson Orin Nano modules---one (A) running the robot-side navigation stack and one (B) hosting the Qwen3.5-9B planner and the retrieval pipeline---so that the planner's skill calls drive navigation and perception while the camera stream updates the food-accessibility store.

Nutri-ATLAS couples the nutrition pipeline with an LLM-guided robotic skill interface (Fig.~\ref{fig:system_overview}(b)), letting the agent actively acquire missing evidence from the physical environment - for instance, checking whether a recommended ingredient is present in the kitchen before finalizing a suggestion. The LLM planner interacts with the robot through a unified skill library spanning navigation, perception, preference management, and recommendation:
\begin{itemize}
  \item \texttt{list\_landmarks}: returns known semantic landmarks and their map coordinates;
  \item \texttt{navigate\_to\_landmark}: moves the robot to a named landmark or object location;
  \item \texttt{get\_detected\_objects}: queries the persistent object map accumulated during the session;
  \item \texttt{update\_preference}: records dietary preferences and constraints into the Preference Memory;
  \item \texttt{update\_food\_access}: updates the Food Accessibility store with newly observed food items;
  \item \texttt{food\_recommandation}: invokes the four-stage RAG pipeline (Sec.~\ref{subsec:recommendation}), conditioning Stage~4 meal composition on the current Food Accessibility state.
\end{itemize}

The agent maintains three persistent stores (Fig.~\ref{fig:system_overview}(b)): a \textit{Landmark Database} of room-level locations, a \textit{Preference Memory} of dietary restrictions and prior choices, and a \textit{Food Accessibility} store of items confirmed present. The Food Accessibility store is passed to Stage~4 as an availability signal, grounding recipe suggestions in foods the agent has actually observed. Given a request, the planner first checks whether the current knowledge base suffices; if not, it issues a sequence of skill calls to gather evidence before responding. For ``Suggest a high-protein snack I can grab from the kitchen,'' the agent lists landmarks, navigates to the kitchen, queries detected objects, updates food accessibility, and invokes \texttt{food\_recommandation} with the updated context - the closed-loop workflow of Fig.~\ref{fig:high_level_system_overview}. Detections are filtered for relevance and accumulated across the session,
\begin{equation}
\mathcal{K}_{t+1}=\mathcal{K}_{t}\cup\{\mathrm{LLM}_{\mathrm{filter}}(\mathcal{D}_t)\},
\end{equation}
where $\mathcal{D}_t$ denotes detections at step $t$, letting the agent build environmental knowledge over long-horizon interactions.

We assess the LLM planner's ability to select and sequence embodied skills on EmbodiedBench-ALFRED suite~\cite{yang2025embodiedbench}, a standard benchmark of long-horizon embodied tasks. To mirror Nutri-ATLAS's use of persistent environmental memory - the landmark and food-accessibility stores it accumulates during operation - we equip the planner with an object memory that carries observations across steps and surfaces the relevant ones at each decision. We report task success rate across EmbodiedBench-ALFRED subsets. Beyond simulation, we further validate the complete pipeline through a real-world deployment, where the planner executes the embodied skills end-to-end to produce grounded nutrition recommendations.

\section{Nutri-ATLAS Evaluation Setup} \label{sec:evaluation_setup}


We evaluate Nutri-ATLAS along three complementary dimensions: 
(i) offline nutrition reasoning, which measures whether the system can learn nutrition-aware food representations, retrieve reasonable substitutes, ground food mentions, and generate personalized food recommendations; 
(ii) simulated embodied evaluation, which tests the LLM planner's ability to select and execute embodied skills for long-horizon tasks; and 
(iii) real-world robot evaluation, which examines the feasibility of deploying the complete system in a physical home-like environment.

\subsection{Datasets and Knowledge Bases}

\noindent\textbf{USDA FoodData Central and Nutri-ATLAS knowledge base.}
We construct the unified nutrition knowledge base from USDA FoodData Central, combining Foundation Foods (74{,}175 raw entries, cleaned to 4{,}616 after removing isolated laboratory samples) and SR Legacy (7{,}793 curated entries). After embedding-based deduplication, the resulting DuckDB knowledge base contains 9{,}991 canonical food entries, 626 nutrient entries, and the corresponding Food$\rightarrow$Nutrient measurements. FoodKG~\cite{haussmann2019foodkg} contributes an additional 82{,}238 recipes and 238 cuisine/meal-type tags, integrated into the same knowledge base via 752{,}951 recipe-ingredient links.

\noindent\textbf{Heterogeneous nutrition graph.}
The graph $G=(V,E)$ contains four node types: 9{,}991 food, 626 nutrient, 82{,}238 recipe, and 238 tag nodes ($|V|{=}93{,}093$). Edges include Food$\leftrightarrow$Nutrient with $\log(1{+}\mathrm{amount})$ as edge attribute, Recipe$\leftrightarrow$Food (ingredient containment), Recipe$\leftrightarrow$Tag, and substitution edges induced from HealthyFoodSubs. Supervision comes from the Food$\rightarrow$Nutrient relations - split 85/7/8 into train, validation, and test edges - and the substitution edges; recipe and tag embeddings are learned implicitly through joint message passing in the shared 64-dimensional space.

\noindent\textbf{HealthyFoodSubs.}
HealthyFoodSubs~\cite{loesch2024automated} provides human-annotated food substitution pairs over the USDA SR Legacy subset. Given a source food, the task is to rank candidate substitutes such that ground-truth substitutes are ranked highly within the same FdGrp food category. We use the released held-out test split, with each query scored against the within-category candidate pool following the protocol of~\cite{loesch2024automated}. This benchmark evaluates whether the learned food representations capture substitution relationships beyond text similarity alone.

\noindent\textbf{NutriBench~v2.}
We use NutriBench~v2~\cite{dhaliwal2025nutribench}, a benchmark of natural-language meal descriptions paired with ground-truth macronutrient and energy values. The v2 release contains 15,617 meal-description samples spanning 24 countries, with each sample annotated with carbohydrate, protein, fat, and energy values, together with country and serving-type metadata. For each meal, the system extracts food mentions, retrieves candidate USDA entries, injects per-100g nutrient references into the LLM prompt, and predicts total carbohydrate, protein, fat, and energy. Predictions are scored against the per-nutrient tolerance bands defined by~\cite{dhaliwal2025nutribench}.

\noindent\textbf{PFoodReQ.}
We use the PFoodReQ~\cite{chen2021personalized} test set of 2{,}269 queries (1{,}542 in-domain and 727 out-of-domain). Each query is associated with a natural-language question, persona-derived ingredient preferences and dislikes, optional nutrient guidelines, and on average 2.7 ground-truth recipe answers under approximately 5 constraints. Candidate recipes are drawn from the FoodKG-derived corpus of 82{,}238 recipes shared with the knowledge base.

\noindent\textbf{Clinical patient profiles.}
For the integrated case study (Sec.~\ref{sec:integrated_system_results}), we use a public Diet Recommendation Dataset~\cite{ziya_diet_dataset} of 1{,}000 patient profiles. Each profile lists demographics and vitals, a chronic condition (diabetes, hypertension, obesity, or none), daily caloric intake, a dietary restriction (low-sodium, low-sugar, or none), and a food allergy (peanuts, gluten, or none). The dataset's target-diet label is fully determined by the chronic condition (diabetes\,$\rightarrow$\, low-carb, hypertension\,$\rightarrow$\,low-sodium, otherwise balanced); we therefore use the condition to set the clinical macro direction and evaluate whether Nutri-ATLAS's generated meals are constraint-respecting and macro-appropriate, rather than predicting the label.

\noindent\textbf{EmbodiedBench-ALFRED}
For embodied evaluation, we use the EmbodiedBench-ALFRED suite of EmbodiedBench~\cite{yang2025embodiedbench}, a standard benchmark of long-horizon vision-language embodied tasks. It spans six capability subsets - base, common-sense, complex-instruction, long-horizon, spatial, and visual-appearance - with 50 episodes per subset, reporting task success rate. At each step the planner perceives the current observation and selects an action-space skill to progress toward the goal. 
To mirror Nutri-ATLAS's persistent environmental stores, we optionally augment the planner with an observed-object memory that accumulates detections across steps and retrieves the most instruction-relevant entries at each decision. We evaluate two quantized Qwen3.5-9B backbones, IQ2\_M(Config 2) and Q4\_K\_S(Config 7), and compare against open-source MLLM references from the EmbodiedBench leaderboard.

\begin{figure*}[htbp]
  \centering
  \includegraphics[width=\textwidth]{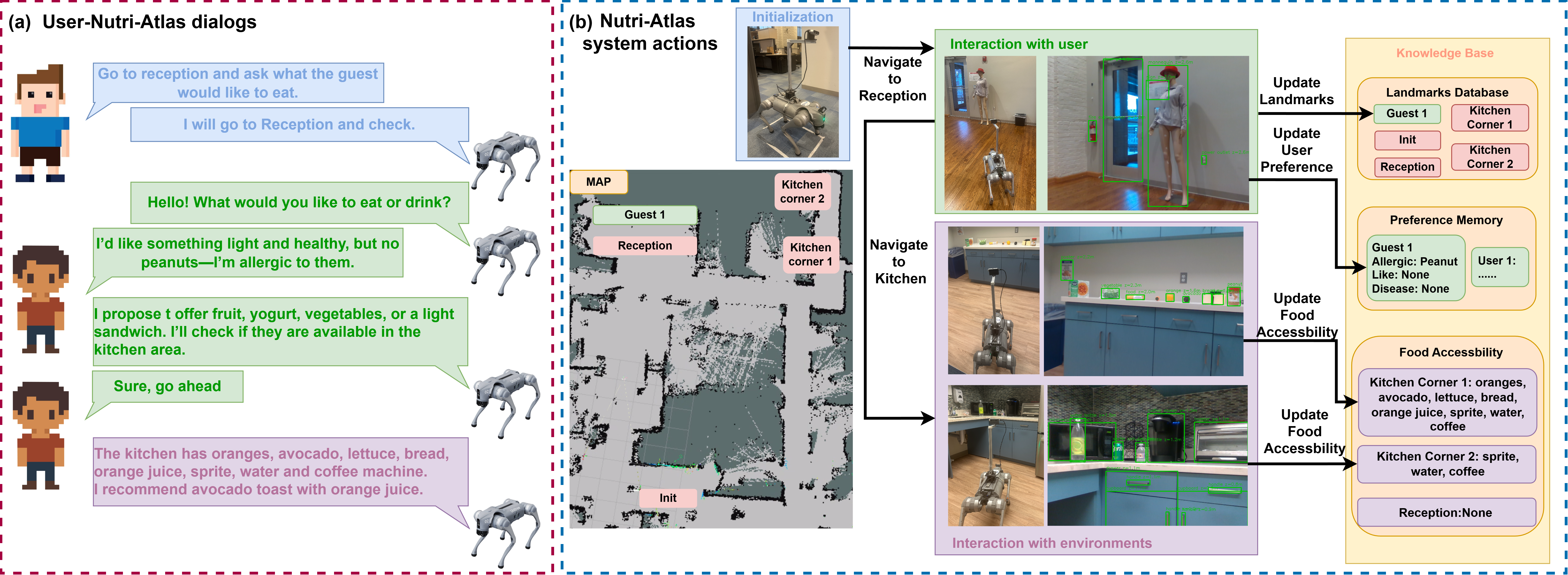}
  \caption{\textbf{Nutri-ATLAS Real-World testing workflow for closed-loop embodied nutrition assistance. }(a) Example user-robot dialogue in which the robot is asked to visit the reception area, query the guest's dietary preference, and return a recommendation that adheres to the stated peanut allergy. The dialogue illustrates how Nutri-ATLAS combines natural-language interaction with embodied information gathering rather than relying only on an initial user prompt. (b) Corresponding embodied system actions executed in the physical environment. Starting from the initial robot pose, the planner uses the mapped environment and landmark database to navigate to the reception area, interact with the guest, update the landmark and preference memories, and then navigate to kitchen locations for food inspection. Vision-based detections are used to update the food-accessibility store at each kitchen region, recording available items such as oranges, avocado, lettuce, bread, orange juice, sprite, water, and coffee. The final recommendation is generated from the updated knowledge base, preference memory, and observed food availability, demonstrating the full loop from user request to navigation, perception, memory update, and grounded nutrition recommendation.}
  \label{fig:realword-test}
\end{figure*}

\subsection{Baselines}

\noindent\textbf{HealthyFoodSubs GAT baseline.}
For food substitution retrieval, we compare against the GAT-based baseline reproduced from Loesch et al.~\cite{loesch2024automated}, trained on food substitution relationships and serving as the published reference for this benchmark.

\noindent\textbf{Nutri-ATLAS GAT, Text, and Hybrid substitution retrieval.}
We evaluate three Nutri-ATLAS substitution variants. \textit{GAT} ranks candidates using only GATv2 food-embedding similarity. \textit{Text} ranks candidates using only Qwen3-Embedding cosine similarity over USDA descriptions. \textit{Hybrid} combines the two via $s(x){=}\alpha\, s_{\text{GAT}}(x){+}(1{-}\alpha)\, s_{\text{text}}(x)$ with $\alpha{=}0.5$.

\noindent\textbf{Dense + GAT retrieval (Nutri-ATLAS RAG).}
Dense + GAT performs dense (text) retrieval of the top-$K$ USDA candidates for each food mention, expands them with GAT-embedding neighbors, and retains the multiple candidates whose cosine similarity exceeds $0.60$, surfacing nutritionally coherent matches that text retrieval alone would miss.

\noindent\textbf{pFoodReQ.}
For personalized recommendation, we compare Nutri-ATLAS against the pFoodReQ baseline (BAMnet)~\cite{chen2021personalized} on the PFoodReQ test set.

\noindent\textbf{Embodied planning baselines.}
On EmbodiedBench-ALFRED, we compare the base planner, which conditions only on the current observation, against the same planner augmented with persistent observed-object memory, and report open-source MLLM results from the EmbodiedBench leaderboard as external references.

\subsection{Metrics}

\noindent\textbf{Food substitution retrieval.}
On HealthyFoodSubs we report Mean Average Precision (MAP) and Recall Ratio at $k$ (RR@$k$ for $k{=}5,10$). MAP measures overall ranking quality, while RR@$k$ measures the fraction of queries whose ground-truth substitute appears in the top-$k$ candidates within the same food-category cohort.

\noindent\textbf{Food--nutrient graph prediction.}
We evaluate the GAT encoder on the held-out validation set with existence AUC for Food$\rightarrow$Nutrient edge prediction and Root-Mean-Square Error (RMSE) for nutrient amount regression in $\log(1{+}\mathrm{amount})$ space; the validation RMSE also drives learning-rate scheduling.


\noindent\textbf{Nutrition estimation.}
On NutriBench~v2 we report per-nutrient classification accuracy under the published tolerance bands and Mean Absolute Error (MAE), reported in grams for carbohydrate, protein, and fat, and in kilocalories for energy. The metrics are used to compare quantized Qwen3.5-9B variants (Q2--Q4).

\noindent\textbf{Recommendation quality.}
On PFoodReQ we report Mean Average Precision (MAP), Mean Average Recall (MAR), and macro-averaged F1, evaluated against the published ground-truth answer sets per query.

\noindent\textbf{Clinical recommendation case study.} 
For the patient case study (Table~\ref{tab:patient_casestudy}) we report, per patient, allergen safety (the recommended recipe contains none of the patient's allergens, checked against ingredient and recipe names) and carbohydrate adherence against published references: a meal is \emph{Low-Carb} when carbohydrate supplies under 26\% of its energy~\cite{feinman2015dietary} and \emph{Balanced} when carbohydrate lies within the AMDR 45--65\%~\cite{standing2005dietary}, together with the recipe's per-serving energy and macronutrients.

\noindent\textbf{Embodied planning.}
On EmbodiedBench-ALFRED we report task success rate - the fraction of episodes in which the agent completes the requested goal - per capability subset, averaged over 50 episodes per subset.

\noindent\textbf{Real-world deployment.}
For real-world experiments we additionally report navigation success, object inspection success, end-to-end completion rate, and representative failure modes, such as missed detections, navigation failures, or incomplete evidence acquisition.

\subsection{Implementation Details}

\noindent\textbf{LLM and embedding backbones.}
The main LLM backbone is Qwen3.5-9B with 4-bit quantization. For the quantization study, we evaluate nine Q2-Q4 variants ranging from Config 1 ($\sim$3.0 GB) to Config 9 ($\sim$6.0 GB) under the fixed Dense + GAT retrieval setting, all 9 configurations are deployable on the Jetson Orin Nano. Dense retrieval uses Qwen3-Embedding-0.6B (28-layer decoder transformer, last-token pooling, 1{,}024-dimensional output) to encode food mentions, USDA descriptions, and recipe names into a shared embedding space; sparse retrieval is implemented with DuckDB full-text search.

\noindent\textbf{Graph encoder.}
The graph encoder is a 2-layer GATv2 with edge-feature attention. Layer~1 uses 4 attention heads with per-head hidden dimension $d_h{=}64$ (concatenated to $\mathbb{R}^{N\times 256}$); layer~2 reduces to a single head with $d_h{=}64$, producing $\mathbf{h}_2{\in}\mathbb{R}^{N\times 64}$. Each layer applies ELU activation, dropout $p{=}0.2$, a residual linear projection (from the previous-layer width to the current-layer width), and LayerNorm. Initial node features are the sum of a learned node embedding and a 4-way type embedding (food, nutrient, recipe, tag).

\noindent\textbf{GAT training objective.}
The model is optimized jointly on three tasks: nutrient-edge existence prediction, nutrient amount regression, and food-substitution prediction:
\begin{equation}
    \mathcal{L} = \mathcal{L}_{amt} + 0.4\,\mathcal{L}_{exist} + \mathcal{L}_{sub},
\end{equation}
where $\mathcal{L}_{amt}$ is SmoothL1 over standardized $\log(1{+}\mathrm{amount})$ targets, and $\mathcal{L}_{exist}$, $\mathcal{L}_{sub}$ are binary cross-entropy losses with sampled non-edges as negatives. We train for 90 epochs with the AdamW optimizer (learning rate $2{\times}10^{-3}$, weight decay $1{\times}10^{-4}$), and validation RMSE drives a ReduceLROnPlateau schedule (factor 0.6, patience 4). Recipe and tag node embeddings receive no direct loss term but are shaped indirectly through message passing into the supervised food nodes.

\noindent\textbf{Retrieval configuration.}
For nutrition estimation, food terms extracted from each NutriBench meal are matched to USDA entries using Dense+GAT retrieval---dense top-$K$ retrieval expanded by GAT-embedding neighbors and filtered to multi-candidate matches above a $0.60$ cosine-similarity threshold; the retrieved per-100\,g nutrient values are inserted into the LLM prompt as external references. For food substitution and recipe retrieval, candidate items are ranked using GAT similarity, text similarity, or the hybrid score $s{=}0.5\,s_{\text{GAT}}{+}0.5\,s_{\text{text}}$.

\noindent\textbf{Hardware.}
GAT training is conducted on a single NVIDIA RTX A6000 GPU. Offline evaluation, including NutriBench, PFoodReQ, and HealthyFoodSubs, runs on a workstation with the LLM served by llama.cpp on the same GPU. Real-world embodied experiments use a Unitree Go2 robot with an Intel RealSense RGB-D camera, with perception and planning executing on the operator workstation. For the hardware-aware analysis (Sec.~\ref{sec:model_selection_results}), the GGUF-quantized variants are additionally profiled on an NVIDIA Jetson Orin Nano to obtain the throughput and energy-per-token measurements.

\begin{table*}[!t]
\centering
\begin{tabular}{|l|c|c|c|}
\hline
\textbf{Method} & \textbf{MAP (\%)} & \textbf{RR@5 (\%)} & \textbf{RR@10 (\%)} \\
\hline
RDF2Vec~\cite{loesch2022automated}
& $15.4$ & $35.9$ & $43.8$ \\
\hline
GraphSAGE~\cite{loesch2024automated}
& $27.4$ & $57.7$ & $69.3$ \\
\hline
HealthyFoodSubs GAT~\cite{loesch2024automated}
& $34.5$ & $68.0$ & $75.7$ \\
\hline
Nutri-ATLAS GATv2
& $33.2$ & $73.5$ & $81.2$ \\
\hline
Nutri-ATLAS Text
& $18.6$ & $52.5$ & $68.0$ \\
\hline
\textbf{Nutri-ATLAS Hybrid (ours)}
& $\mathbf{37.9}$ & $\mathbf{80.7}$ & $\mathbf{90.1}$ \\
\hline
\end{tabular}
\caption{Nutri-ATLAS food substitution retrieval evaluation on HealthyFoodSubs~\cite{loesch2024automated}.
MAP, RR@5, and RR@10 are reported as percentages, under the within-category (filtered-ranking) protocol of Loesch et al.~\cite{loesch2024automated}.
RDF2Vec~\cite{loesch2022automated}, GraphSAGE, and GAT are reproduced from Loesch et al.~\cite{loesch2022automated,loesch2024automated}.
Nutri-ATLAS GATv2 ranks substitutes using only GATv2 food-embedding similarity, Nutri-ATLAS Text uses only text-embedding similarity, and Nutri-ATLAS Hybrid combines graph-based nutritional similarity with text-based semantic similarity.
The best result in each column is highlighted in bold.}
\label{tab:food_substitution_results}
\vspace{-8pt}
\end{table*}

\begin{table*}[t]
\centering
\footnotesize
\begin{tabular}{|p{1.8cm}|p{2.5cm}|c|p{3.0cm}|c|l|}
\hline
\textbf{Patient} & \textbf{Restriction (disease $\cdot$ allergy)} & \textbf{Require\-ment} & \textbf{Nutri-ATLAS Recommended Meal} & \textbf{Nutrition (kcal; P/C/F g)} & \textbf{Requirement met} \\
\hline
P0479, 77M, BMI 22.1 & Diabetes $\cdot$ peanut allergy & Low-Carb & \textit{Stuffed Hamburger-Cabbage Buns} & 393; 30/5/24 & \makecell[l]{Low-Carb 6\,\%E ($<$26) \checkmark \\ peanut-free \checkmark} \\ \hline
P0628, 20F, BMI 26.8 & Diabetes $\cdot$ gluten allergy & Low-Carb & \textit{Easy Jalapeno Cilantro Chicken Salad} & 287; 30/2/14 & \makecell[l]{Low-Carb 4\,\%E ($<$26) \checkmark \\ gluten-free \checkmark} \\ \hline
P0053, 69M, BMI 40.0 & Obesity $\cdot$ peanut allergy & Balanced & \textit{Caldo Verde} & 407; 14/50/15 & \makecell[l]{Balanced 52\,\%E (45--65) \checkmark \\ peanut-free \checkmark} \\ \hline
P0389, 69M, BMI 27.1 & Obesity $\cdot$ gluten allergy & Balanced & \textit{Tangy Potato Salad} & 385; 6/60/14 & \makecell[l]{Balanced 62\,\%E (45--65) \checkmark \\ gluten-free \checkmark} \\ \hline
P0605, 70F, BMI 36.2 & None $\cdot$ no allergy & Balanced & \textit{Easy Hummus} & 292; 14/42/8 & \makecell[l]{Balanced 58\,\%E (45--65) \checkmark} \\ \hline

\end{tabular}
\caption{Personalized clinical meal recommendation on five patients spanning the Low-Carb (diabetes) and Balanced (obesity/none) targets and both allergens. Requirements are measured against published references: \textbf{Low-Carb} = carbohydrate $<$26\% of meal energy~\cite{feinman2015dietary}; \textbf{Balanced} = carbohydrate within the AMDR 45--65\% of meal energy~\cite{standing2005dietary}; \textbf{allergen-free} verified from ingredient names against the major allergens designated by FALCPA~\cite{thompson2006food}. Energy is derived from the macros (4/4/9). Macros are per serving; the corpus stores no per-recipe sodium or sugar, so those are not claimed.}
\label{tab:patient_casestudy}
\end{table*}

\subsection{Robot Platform and Environment Setup}

The embodied version of Nutri-ATLAS is deployed on a Unitree Go2 quadruped equipped with an Intel RealSense RGB-D camera, with an open-vocabulary VLM grounding fallback). The robot maintains three persistent stores: a \textbf{Landmark Database} of 4 fixed room-level locations (kitchen, living room, bedroom, bathroom) with their map coordinates, a \textbf{Preference Memory} of dietary restrictions and prior choices, and a \textbf{Food Accessibility} store that accumulates food items observed during the session and is cleared at session start.

The LLM planner interacts with the robot through a typed skill interface rather than free-form text parsing. The available skills are:
\begin{itemize}
    \item \texttt{list\_landmarks}: returns all known landmarks and their coordinates;
    \item \texttt{navigate\_to\_landmark}: moves the robot to a named landmark or coordinate;
    \item \texttt{get\_detected\_objects}: returns the persistent object map accumulated during the session;
    \item \texttt{update\_preference}: records dietary preferences and constraints into the Preference Memory;
    \item \texttt{update\_food\_access}: updates the Food Accessibility store with newly observed food items;
    \item \texttt{food\_recommandation}: invokes the four-stage retrieval pipeline (Sec.~\ref{subsec:recommendation}), conditioning meal composition on the current Food Accessibility state.
\end{itemize}

The embodied system is exercised in two settings: the planner's skill selection is benchmarked in simulation on EmbodiedBench-ALFRED, while the full closed loop is deployed on the physical robot in an indoor home-like environment with kitchen and living-room regions. Figure~\ref{fig:realword-test} traces one real-world episode. The user requests a light, healthy meal and reports a peanut allergy; rather than replying from the request alone, the planner visits the reception and kitchen, inspects the foods available through the RealSense stream, and writes the new observations into its Preference Memory and Food Accessibility stores. It then composes the meal over only the foods actually present and returns an allergen-safe recommendation grounded in the observed environment.

\section{Nutri-ATLAS Evaluation Results}
\label{sec:results}

We present results for the evaluation of Nutri-ATLAS: graph-based food representation learning (Sec.~\ref{sec:graph_kb_results}), personalized nutrition-aware retrieval and recommendation (Sec.~\ref{sec:recommendation_results}), hardware-aware model selection (Sec.~\ref{sec:model_selection_results}), and the fully integrated robotic system (Sec.~\ref{sec:integrated_system_results}). Datasets, baselines, metrics, and implementation details are described in Sec.~\ref{sec:evaluation_setup}.

\begin{table*}[!t]
\centering
\begin{tabular}{|l|c|c|c|c|c|c|c|c|}
\hline
\textbf{Model} & \multicolumn{2}{c|}{\textbf{Carbohydrate}} & \multicolumn{2}{c|}{\textbf{Protein}} & \multicolumn{2}{c|}{\textbf{Fat}} & \multicolumn{2}{c|}{\textbf{Energy}} \\
\hline
 & Accuracy (\%) & MAE & Accuracy (\%) & MAE & Accuracy (\%) & MAE & Accuracy (\%) & MAE \\
\hline
Config 1: Qwen3.5-9B-UD-IQ2\_XXS 
& $24.6$ & $85.29$ 
& $50.7$ & $29.25$ 
& $49.5$ & $24.71$ 
& $25.5$ & $567.14$ \\
\hline
Config 2:Qwen3.5-9B-UD-IQ2\_M
& $41.8$ & $46.85$ 
& $71.8$ & $13.96$ 
& $63.1$ & $15.55$ 
& $35.5$ & $213.45$ \\
\hline
Config 3:Qwen3.5-9B-UD-Q2\_K\_XL 
& $40.1$ & $67.82$ 
& $71.5$ & $47.11$ 
& $63.6$ & $21.68$ 
& $44.2$ & $190.18$ \\
\hline
Config 4:Qwen3.5-9B-Q3\_K\_S 
& $46.1$ & $71.18$ 
& $73.0$ & $32.54$ 
& $63.6$ & $18.66$ 
& $47.2$ & $337.15$ \\
\hline
Config 5:Qwen3.5-9B-Q3\_K\_M 
& $52.2$ & $17.57$ 
& $\mathbf{79.3}$ & $\mathbf{5.82}$ 
& $67.2$ & $9.16$ 
& $\mathbf{51.2}$ & $\mathbf{121.24}$ \\
\hline
Config 6:Qwen3.5-9B-UD-Q3\_K\_XL 
& $51.7$ & $20.24$ 
& $78.5$ & $6.51$ 
& $67.4$ & $12.33$ 
& $50.0$ & $139.06$ \\
\hline
Config 7:Qwen3.5-9B-Q4\_K\_S
& $52.9$ & $16.89$ 
& $78.0$ & $6.12$ 
& $68.0$ & $\mathbf{8.99}$ 
& $51.1$ & $135.53$ \\
\hline
Config 8:Qwen3.5-9B-Q4\_K\_M
& $52.8$ & $19.29$ 
& $78.5$ & $8.30$ 
& $\mathbf{68.6}$ & $9.71$ 
& $49.9$ & $126.55$ \\
\hline
Config 9:Qwen3.5-9B-UD-Q4\_K\_XL 
& $\mathbf{54.0}$ & $\mathbf{16.51}$ 
& $78.6$ & $8.40$ 
& $67.9$ & $11.19$ 
& $\mathbf{51.2}$ & $126.26$ \\
\hline
\end{tabular}
\caption{Performance comparison of quantized Qwen3.5-9B variants on NutriBench~v2 under Nutri-ATLAS's fixed Dense + GAT retrieval setting. Accuracy denotes the fraction of predictions within the predefined tolerance of the ground truth. MAE is reported in grams for macronutrients and kcal for energy. Models are grouped by quantization level from Q2 to Q4 and ordered from smaller to larger variants within each group. The best result in each column is highlighted in bold.}
\label{tab:quantized_model_results}
\end{table*}

\subsection{Unified Nutrition Graph Knowledge Base Results}
\label{sec:graph_kb_results}

We first evaluate whether the unified nutrition graph learns food representations useful beyond nutrient prediction, using the HealthyFoodSubs substitution task: ranking candidate substitutes for a given source food. We compare three Nutri-ATLAS retrieval variants - (i) \textit{GAT}, ranking by GATv2 food-embedding similarity only; (ii) \textit{Text}, ranking by text-embedding similarity only; and (iii) \textit{Hybrid}, combining graph-based nutritional similarity with text-based semantic similarity - and report the published baseline of Loesch et al.~\cite{loesch2024automated} for reference.

As shown in Table~\ref{tab:food_substitution_results}, Nutri-ATLAS GATv2 matches the published baseline on MAP (33.2\% vs.\ 34.5\%) while improving top-$k$ recall, raising RR@5 from 68.0\% to 73.5\% and RR@10 from 75.7\% to 81.2\%. The gains in top-$k$ recall indicate stronger coverage of relevant substitutes, which is especially useful when multiple substitutes can be recommended. The text-only variant performs worse than the graph-based variant, showing that semantic similarity alone is insufficient for substitution search, while the hybrid variant is strongest overall (37.9\% MAP, 80.7\% RR@5, 90.1\% RR@10) and surpasses the baseline on all three metrics, confirming that nutritional graph structure and text similarity provide complementary signals. Unlike the baseline, Nutri-ATLAS trains GATv2 over a broader heterogeneous graph with nutrient-amount regression, nutrient-edge prediction, and substitution prediction, so the same embeddings support both substitute retrieval and downstream nutrition-aware recommendation.


\begin{table}[!t]
\centering
\begin{tabular}{lccc}
\hline
Method & MAP (\%) & MAR (\%) & F1 \\
\hline
\textsc{pFoodReQ}~\cite{chen2021personalized} & 62.7 & 61.8 & 63.7 \\
\hline
\textbf{Nutri-ATLAS Food Recommendation} & \textbf{78.8} & \textbf{83.0} & \textbf{77.5} \\
\hline
\end{tabular}
\caption{Nutri-ATLAS personalized food recommendation results on the PFoodReQ~\cite{chen2021personalized} test set.}
\label{tab:pfoodreq_results}
\vspace{-10pt}
\end{table}

\subsection{Personalized and Clinical Food Recommendation Results}
\label{sec:recommendation_results}

We evaluate Nutri-ATLAS's personalized, graph-aware recommendation on two complementary sets: a clinical patient case study, which probes per-patient constraint adherence (carbohydrate target and allergen safety), and the PFoodReQ test set, which measures recipe-recommendation quality against a published baseline.

\subsubsection{Clinical Meal Recommendation}
\label{sec:clinical_results}

We test whether Nutri-ATLAS's recommendations satisfy individual clinical constraints on a case study of patient profiles from a public diet-recommendation dataset~\cite{ziya_diet_dataset}, spanning the two carbohydrate-grounded targets the recipe corpus lets us measure---\emph{Low-Carb} (diabetes) and \emph{Balanced} (obesity or no condition)---crossed with peanut and gluten allergies. For each patient, the LLM gap analysis (Sec.~\ref{subsec:recommendation}) sets a condition-appropriate macro direction and the retrieval stages return a concrete recipe that meets it while hard-excluding the patient's allergens. Table~\ref{tab:patient_casestudy} reports representative cases, each verified against published references: a meal is \emph{Low-Carb} when carbohydrate supplies under 26\% of its energy~\cite{feinman2015dietary} and \emph{Balanced} when carbohydrate falls within the AMDR 45--65\%~\cite{standing2005dietary}, and allergen-safe when no ingredient matches the patient's allergen terms~\cite{thompson2006food}. Across the five cases the diabetic patients receive low-carbohydrate meals (4--6\% of energy) and the others moderate-carbohydrate meals (52--62\%), while every peanut- or gluten-allergic patient receives an allergen-free recipe. Because neither the recipe corpus nor the patient records store sodium or sugar, we restrict the measured claims to carbohydrate direction and allergen safety rather than asserting low-sodium or low-sugar adherence.

\begin{figure}[!b]
  \centering
  \includegraphics[width=\columnwidth]{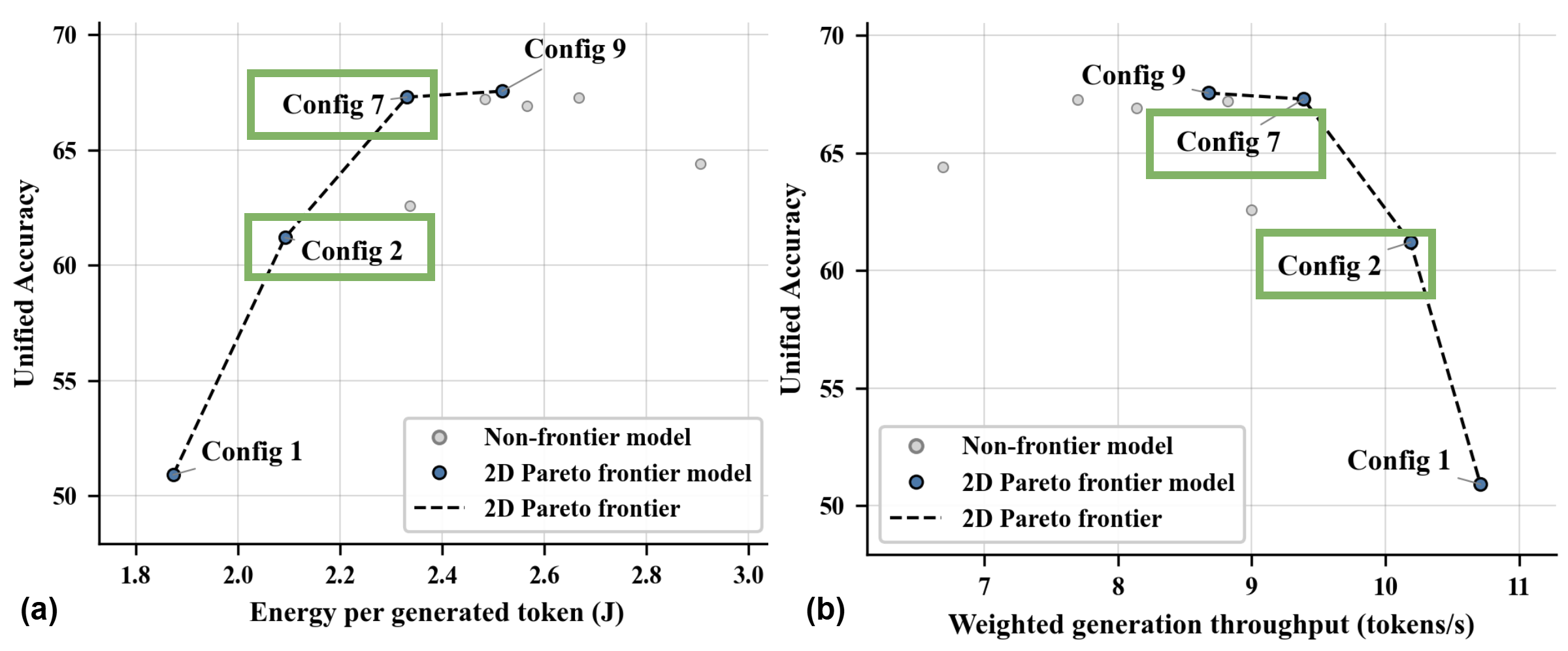}
    \caption{\textbf{Pareto frontier analysis for quantized Qwen3.5-9B configurations in Nutri-ATLAS Throughtput and Energy measured on the Jetson Orin Nano.}
    (a) Unified accuracy Score (calculated using Equation~\ref{eq:unified_acc}) versus energy per generated token, where lower energy is preferred and (b) versus weighted generation throughput, where higher throughput is preferred.
    Blue points denote two-dimensional Pareto frontier models, gray points denote non-frontier models, and dashed lines connect the frontier.}
  \label{fig:pareto_acc_tradeoff}
\end{figure}

\begin{figure}[!b]
\centering
\includegraphics[width=0.9\columnwidth]{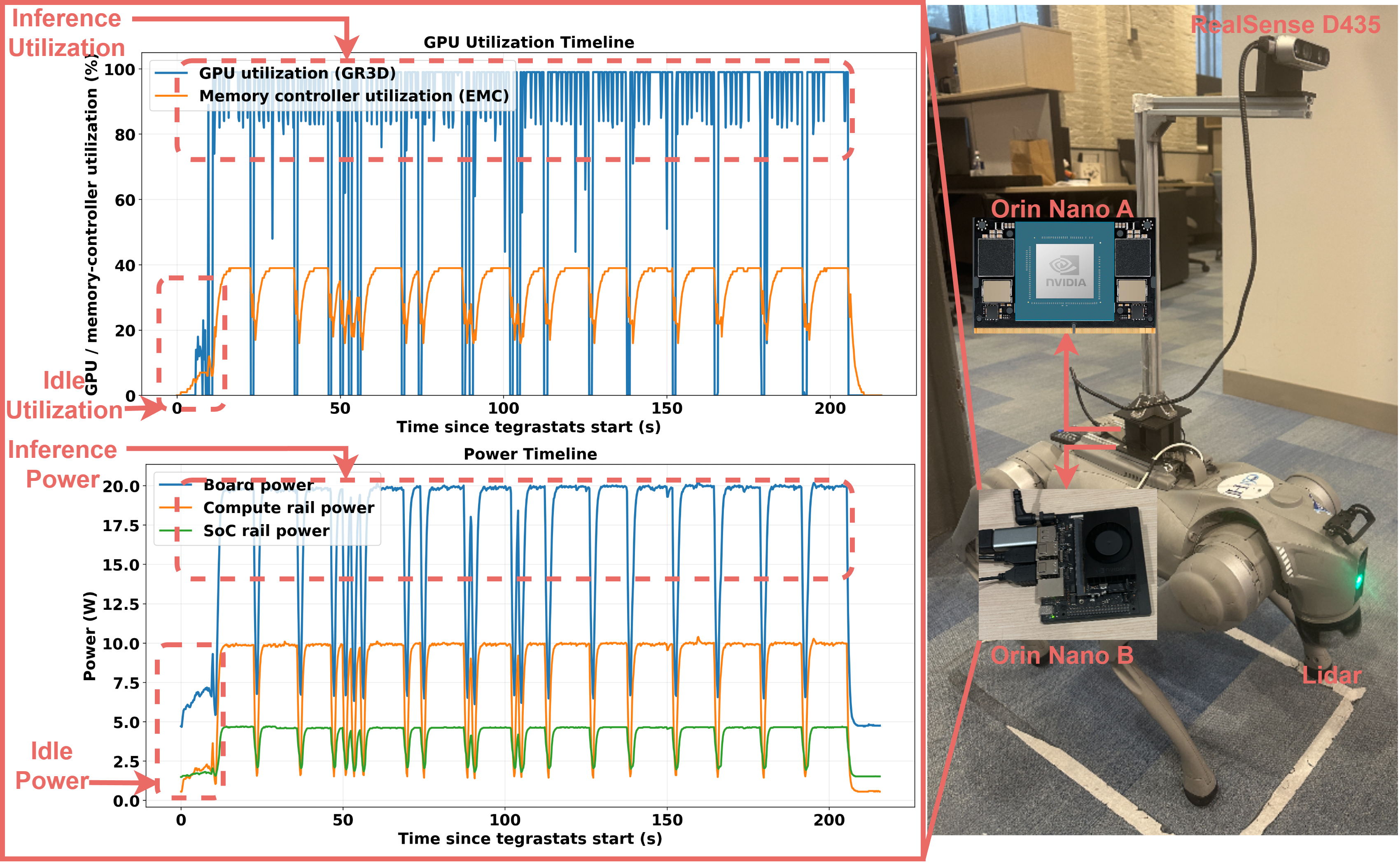}
\caption{\textbf{On-device hardware profiling of the deployed Nutri-ATLAS planner} (Config~2) on the server-side Jetson Orin Nano. \textbf{Top:} GPU (GR3D) and memory-controller (EMC) utilization across a sequence of nutrition-assistant inference requests. \textbf{Bottom:} board, compute-rail, and SoC-rail power. Utilization and power rise sharply during each generation and fall back to an idle floor between requests (idle and inference phases marked), yielding the on-device throughput and energy-per-token used in the hardware-aware model selection.}
\label{fig:hardware_power}
\end{figure}

\begin{figure*}[!t]
  \centering
  \includegraphics[width=\textwidth]{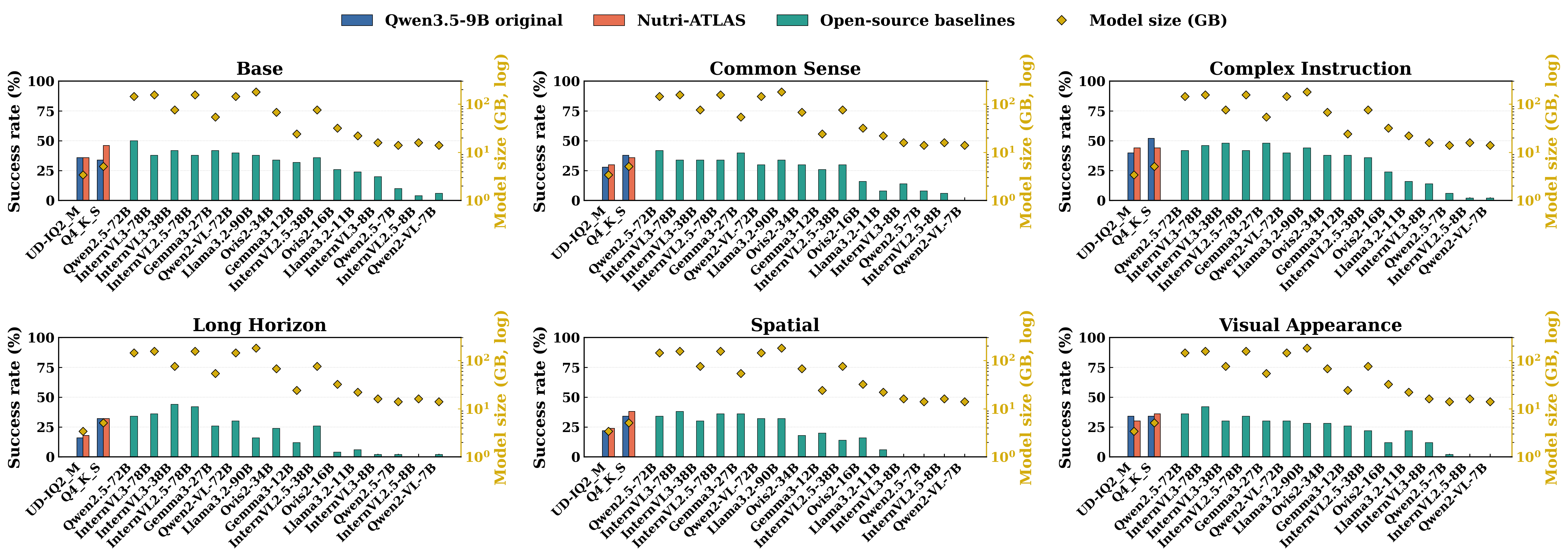}
    \caption{\textbf{Per-subtask success rate of Nutri-ATLAS on EmbodiedBench-ALFRED (50 episodes per subtask).} For our quantized Qwen3.5-9B GGUF runs (Config 2, Config 7), paired bars compare the baseline planner (Origin) against the same model augmented with a persistent observed-object memory (With memory). Single bars are open-source MLLM references from the EmbodiedBench leaderboard. Diamonds on the right log-scale axis show model size (GB): on-disk size for our GGUF runs, FP16 weight footprint for references.}
    \label{fig:eb_alfred_per_subtask}
\end{figure*}

\subsubsection{Personalized Food Recommendation Evaluation}
\label{sec:pfoodreq_results}

Table~\ref{tab:pfoodreq_results} compares Nutri-ATLAS with the pFoodReQ baseline on the PFoodReQ test set. Nutri-ATLAS reaches 78.8\% MAP, 83.0\% MAR, and 77.5\% F1, surpassing the baseline by 16.1, 21.2, and 13.8 percentage points, respectively. By ranking recipes with combined graph and text similarity and enforcing ingredient and nutrient constraints over the clean augmented knowledge base, Nutri-ATLAS surfaces relevant recipes that lexical matching alone would miss.

\subsection{Hardware-Aware LLM selection Results}
\label{sec:model_selection_results}

We next evaluate the different quantized Qwen3.5-9B configurations for resource-constrained deployment using NutriBench-v2 nutrient-estimation accuracy together with on-device generation throughput and energy-per-token measurements. Table~\ref{tab:quantized_model_results} reports per-nutrient accuracy for each quantized Qwen3.5-9B variant under the fixed Dense + GAT setting; no single variant dominates all nutrients -- for example, Config 9 leads on carbohydrate while Config 5 leads on protein and energy --- motivating a multi-objective view. Fig.~\ref{fig:pareto_acc_tradeoff} shows that no single configuration is optimal across all objectives: Q4-level models give the strongest accuracy, while smaller Q2-level models offer higher throughput or lower energy. The Pareto frontier therefore provides a practical way to select a model under deployment constraints --- an accuracy-first setting favors a Q4-level model, while an energy-first or performance-first setting favors a smaller Q2-level model.

The strongest scores are achieved by Q4- and Q3-level models, with Config 9 obtaining the highest unified score, while very aggressive quantization degrades sharply. Together, the unified score and Pareto analysis offer complementary criteria: the unified score identifies the strongest nutrition-reasoning model, while the Pareto frontier exposes deployment-specific trade-offs among accuracy, throughput, and energy.

For the embodied deployment, on-robot energy and latency dominate, so we adopt the energy-first Q2-level configuration (Config~2) identified by the Pareto analysis rather than the most accurate Q4-level model. Fig.~\ref{fig:hardware_power} profiles this planner directly on the server-side Jetson Orin Nano, tracing GPU and memory-controller utilization and the board and per-rail power draw across a sequence of nutrition-assistant requests: each generation drives utilization and power from an idle floor up to a sustained inference plateau and back, and integrating power over the generated tokens gives the energy-per-token figures reported above. The trace confirms that the IQ2-level planner sustains interactive generation well within the Orin Nano's power envelope, making it a practical default for the on-robot experiments of Sec.~\ref{sec:model_selection_results}.

\subsection{Fully Integrated Robotic Nutrition System Results}
\label{sec:integrated_system_results}

\begin{table*}[!t]
\centering
\caption{Three-tier real-world embodied nutrition benchmark: the nine evaluation prompts grouped by complexity. \emph{Easy} prompts require nutrition analysis and recommendation; \emph{Medium} add robot navigation and a kitchen availability check that re-plans the meal around observed foods; \emph{Hard} add an allergen or condition constraint.}
\label{tab:realworld_tasks}
\begin{tabular}{|c|c|p{13cm}|}
\hline
\textbf{Complexity} & \textbf{Task ID} & \textbf{Task} \\ \hline
Easy   & \textit{Task 1} & \textit{I had a bowl of oatmeal for breakfast, recommend a lunch.} \\ \cline{2-3}
       & \textit{Task 2} & \textit{I ate a banana and toast for breakfast and a chicken sandwich for lunch. What should I have for dinner?} \\ \cline{2-3}
       & \textit{Task 3} & \textit{I had a chicken salad for breakfast, suggest a lunch and a dinner for me.} \\ \hline
Medium & \textit{Task 4} & \textit{I had two slices of avocado toast for breakfast, please recommend a lunch based on food availability.} \\ \cline{2-3}
       & \textit{Task 5} & \textit{I had yogurt for breakfast and an egg toast for lunch, recommend a light dinner I can make here.} \\ \cline{2-3}
       & \textit{Task 6} & \textit{I had an apple and a cup of milk for breakfast; check the pantry and kitchen to see what I could eat for lunch.} \\ \hline
Hard   & \textit{Task 7} & \textit{I had toast for breakfast. I am allergic to peanuts. Recommend a lunch based on the food available in the kitchen.} \\ \cline{2-3}
       & \textit{Task 8} & \textit{I had an orange for breakfast. I am allergic to peanuts and I have diabetes. Recommend a lunch based on what we have in the kitchen.} \\ \cline{2-3}
       & \textit{Task 9} & \textit{I had yogurt with granola for breakfast. I am allergic to peanuts and gluten. Recommend a lunch by checking the kitchen and pantry.} \\ \hline
\end{tabular}
\end{table*}

\begin{table*}[!t]
\centering
\footnotesize
\caption{Per-task grading of the deployed Nutri-ATLAS planner (Config~2) on the three-tier benchmark (Table~\ref{tab:realworld_tasks}). Each cell reports passed/applicable checkpoints for that subtask; ``0/0'' marks a subtask the task does not exercise. \textbf{Cumulative Success} aggregates all of the task's checkpoints.}
\label{tab:realworld_results}
\begin{tabular}{|l|l|c|c|c|c|c|c|}
\hline
\textbf{Complexity} & \textbf{Task ID} & \makecell{\textbf{Nutrition}\\\textbf{Extraction}} & \makecell{\textbf{Recommend.}\\\textbf{Success}} & \makecell{\textbf{Navigation}\\\textbf{Success}} & \makecell{\textbf{Availability}\\\textbf{Check}} & \makecell{\textbf{Constraint}\\\textbf{Check}} & \makecell{\textbf{Cumulative}\\\textbf{Success}} \\
\hline
Easy & \textit{Task 1} & 1/1 & 1/1 & 0/0 & 0/0 & 0/0 & 2/2 \\
\cline{2-8}
 & \textit{Task 2} & 2/2 & 1/1 & 0/0 & 0/0 & 0/0 & 3/3 \\
\cline{2-8}
 & \textit{Task 3} & 1/1 & 1/2 & 0/0 & 0/0 & 0/0 & 2/3 \\
\hline
Medium & \textit{Task 4} & 1/1 & 1/1 & 1/1 & 3/5 & 0/0 & 6/8 \\
\cline{2-8}
 & \textit{Task 5} & 2/2 & 1/1 & 1/1 & 3/5 & 0/0 & 7/9 \\
\cline{2-8}
 & \textit{Task 6} & 2/2 & 1/1 & 2/2 & 6/8 & 0/0 & 11/13 \\
\hline
Hard & \textit{Task 7} & 1/1 & 1/1 & 1/1 & 5/7 & 1/1 & 9/11 \\
\cline{2-8}
 & \textit{Task 8} & 1/1 & 1/1 & 1/1 & 5/7 & 2/2 & 10/12 \\
\cline{2-8}
 & \textit{Task 9} & 2/2 & 1/1 & 1/2 & 6/10 & 2/2 & 12/17 \\
\hline
\end{tabular}
\end{table*}

Finally, we evaluate Nutri-ATLAS as a complete embodied nutrition assistant in the real world, connecting food grounding, nutritional gap analysis, preference-aware recommendation, robot navigation, scene inspection, and object detection within a single closed-loop workflow. Having established the recommendation pipeline's accuracy and constraint adherence in Sec.~\ref{sec:pfoodreq_results}, we focus here on embodied execution: simulated skill selection and real-world deployment.

We first assess the planner's embodied skill execution on EmbodiedBench-ALFRED (Fig.~\ref{fig:eb_alfred_per_subtask}), reporting per-subtask success over 50 episodes per subtask. For two quantized Qwen3.5-9B backbones (Config 2 and Config 7), we compare the base planner against the same planner augmented with a persistent observed-object memory that mirrors Nutri-ATLAS's environmental stores. The memory-augmented planner improves success on subtasks requiring recall of previously observed objects, while remaining comparable on single-observation subtasks, indicating that accumulated environmental evidence aids long-horizon embodied planning. Despite aggressive quantization and a far smaller on-disk footprint, the Nutri-ATLAS planners remain competitive with the larger open-source MLLM references from the EmbodiedBench leaderboard.

We then deploy the complete system on a physical Unitree Go2 in a home-like environment and evaluate it on a three-tier benchmark of nine natural-language requests (Table~\ref{tab:realworld_tasks}). \emph{Easy} tasks require only nutrition analysis and recommendation; \emph{Medium} tasks add navigation and an availability check that re-plans the meal around the foods the robot finds in the kitchen; and \emph{Hard} tasks further add an allergen or condition constraint. Each request is graded by its constituent subtasks: Table~\ref{tab:realworld_results} reports the passed/applicable checkpoints per subtask and the cumulative per-task success. Running on the planner with Config 2, Nutri-ATLAS handles the Easy tier near-perfectly and degrades gracefully on the harder tiers, where the availability check---mapping live detections onto corpus foods and grounding the refined recommendation---is the dominant failure mode, while nutrition analysis, recommendation, and navigation remain reliable.

In a representative end-to-end interaction (Fig.~\ref{fig:realword-test}), the user reports a previous meal and asks for a recommendation for the next one. Nutri-ATLAS first grounds the reported intake, retrieves nutrient profiles, identifies the nutritional gap, and generates candidate foods (Fig.~\ref{fig:realword-test}(a)). When asked to check what is actually available, the planner navigates to the reception and kitchen, inspects the foods it observes, and updates its preference and food-accessibility stores (Fig.~\ref{fig:realword-test}(b))---before returning a recommendation grounded in observed availability that adheres to the stated peanut allergy.



\section{Conclusion}
\label{sec:conclusion}

This work presented Nutri-ATLAS, a graph-grounded embodied nutrition assistant that integrates food-nutrient knowledge, hybrid graph-text retrieval, personalized recommendation, hardware-aware LLM selection, and robotic evidence acquisition. By combining GATv2-based nutrition embeddings with semantic text embeddings, Nutri-ATLAS improved food substitution retrieval on HealthyFoodSubs, achieving 37.9\% MAP, 80.7\% RR@5, and 90.1\% RR@10, and substantially improved personalized recipe recommendation on PFoodReQ with 78.8\% MAP, 83.0\% MAR, and 77.5\% F1. On NutriBench~v2, Dense+GAT retrieval grounds nutrient estimation across quantized backbones, demonstrating the importance of grounding LLM reasoning in structured nutritional evidence. The hardware-aware evaluation further showed that quantized Qwen3.5-9B variants can be selected through unified accuracy scoring and Pareto analysis to balance accuracy, throughput, and energy efficiency under edge constraints. Robotic skill execution permits Nutri-ATLAS to localize available foods in the physical environment, feeding observations into a personalized recommendation pipeline, closing the loop between dietary reasoning and what the user has accessible. 
Several directions remain open. The current system localizes food availability and generates grounded recommendations, but lacks a bidirectional feedback channel: the robot cannot explain its reasoning to the user, request clarification, or adapt in real time to observed eating behavior. A natural extension would couple Nutri-ATLAS with a clinical nutrition workflow, in which a dietitian sets individualized macronutrient and micronutrient targets that propagate directly into the retrieval pipeline's constraint filters. In a home-deployed setting, this would allow the system to function as an ambient dietary coach, monitoring intake longitudinally, surfacing trends in nutritional gaps, and flagging deviations from clinician-specified targets for asynchronous review. Integrating speech-based interaction and multimodal meal recognition would further reduce the friction of manual logging, the primary barrier to sustained dietary monitoring identified at the outset of this work.  These extensions would shift Nutri-ATLAS from a localization-and-lookup tool toward a genuinely \textit{personalized} dietary support system operating at the interface of clinical nutrition and embodied AI. 

\section*{References}
\vspace{-4ex}
\bibliographystyle{IEEEtran}
\bibliography{references/my_ref, references/eehpc}

\end{document}